\documentclass{article} 
\usepackage{iclr2016_workshop,times}

\usepackage{url}
\usepackage{dsfont}
\usepackage{algorithm}
\usepackage[noend]{algorithmic}
\usepackage{graphicx}
\renewcommand{\algorithmiccomment}[1]{\bgroup\hfill $\triangleright$ ~#1\egroup}
\usepackage{hyperref}
\title{Online Batch Selection for Faster Training of Neural Networks}

\def\R{\mathds{R}}
\def\E{\mathds{E}}
\def\D{{\mathcal{D}}}
\def\H{\textbf{H}}
\def\A{\cal{A}}
\newcommand{\vc}[1]{\textit{\textbf{#1}}}

\author{Ilya Loshchilov \& Frank Hutter \\
Univesity of Freiburg\\
Freiburg, Germany, \\
\texttt{\{ilya,fh\}@cs.uni-freiburg.de} 
}

\begin{document}

\maketitle

\begin{abstract}
Deep neural networks are commonly trained using stochastic non-convex optimization procedures, which are driven by gradient information estimated on fractions (batches) of the dataset. 
While it is commonly accepted that batch size is an important parameter for offline tuning, the benefits of online selection of batches remain poorly understood.
We investigate online batch selection strategies for two state-of-the-art methods of stochastic gradient-based optimization, AdaDelta and Adam. As the loss function to be minimized for the whole dataset is an aggregation of loss functions of individual  datapoints, intuitively, datapoints with the greatest loss should be considered (selected in a batch) more frequently. However, the limitations of this intuition and the proper control of the selection pressure over time are open questions. We propose a simple strategy where all datapoints are ranked w.r.t. their latest known loss value and the probability to be selected decays exponentially as a function of rank. 
Our experimental results on the MNIST dataset suggest that selecting batches speeds up both AdaDelta and Adam by a factor of about 5.
\end{abstract}

\section{Introduction}

Deep neural networks (DNNs) are currently the best-performing method for many classification problems, such
as object recognition from images~\citep{Krizhevsky2012,Donahue_ICML2014} or speech recognition
from audio data~\citep{Deng13newtypes}. 
The standard method to train DNNs is backpropagation via stochastic gradient descent (SGD) based on fixed-size mini-batches of the data.
This training step is the computational bottleneck of DNNs: on large datasets (where DNNs perform particularly well), it requires several days, even on high-performance GPUs, and any speedups would be of substantial value to the booming field of DNNs.

In light of the computational burden of DNN training, we find it surprising that popular DNN
optimization methods do not implement procedures to adaptively choose which datapoints to 
include in a batch.
Instead, it is common in SGD to fix the batch size and iteratively sweep the algorithm over all datapoints. 
Here, we propose and evaluate online batch selection methods to help SGD focus on the most relevant training datapoints and progress  faster.
More specifically, we sample the datapoints for each batch from a non-uniform probability distribution
that depends on the loss previously computed for each datapoint under consideration. 

We first formally describe stochastic gradient descent and the current state-of-the-art DNN optimization procedures AdaDelta \citep{zeiler2012adadelta} and Adam \citep{kingma2014adam}. Then, we describe our proposed online batch selection procedure and demonstrate that it can substantially speed up both AdaDelta and Adam.

\section{Stochastic Gradient Descent (SGD)}

The training of a DNN with $n$ free parameters can be formulated as the problem of minimizing a function $f: \R^n \rightarrow \R$. 
Following the notation of \citet{hazan2015beyond}, we define a loss function $\psi: \R^n \rightarrow \R$ for each training datapoint;
the distribution of training datapoints then induces a distribution over functions $\D$, and the overall function $f$ we aim to optimize is
the expectation of this distribution:
\begin{eqnarray}
	\label{eq:def}
	f(\vc{x}) := \E_{\psi \sim \D}\left[ \psi(\vc{x}) \right].
	\end{eqnarray}

The commonly used procedure to optimize $f$ is to iteratively adjust $\vc{x}_t$ (the parameter vector at time step $t$) using gradient information obtained on a relatively small batch of size $b$. 
More specifically, at each time step $t$ and for a given $\vc{x}_t \in \R^n$, 
a batch $\left\{ \psi^b_{i=1} \right\} \sim \D^b $ is selected 
 to compute $\nabla f_t(\vc{x}_t)$, 
where $f_t(\vc{x}_t)=\frac{1}{b}\sum_{i=1}^b \psi_i (\vc{x}_t)$. 
The Stochastic Gradient Descent (SGD) procedure then becomes a natural extension of the Gradient Descent (GD) to stochastic optimization of $f$ as follows:

\begin{eqnarray}
	\label{eq:sgd}
	\vc{x}_{t+1} = \vc{x}_t - \eta_t \nabla f_t(\vc{x}_t),
\end{eqnarray}

\noindent{}where $\eta_t$ is a learning rate. 
One would like to use second-order information 

\begin{eqnarray}
	\label{eq:newt}
	\vc{x}_{t+1} = \vc{x}_t - \eta_t \vc{H}_t^{-1} \nabla f_t(\vc{x}_t),
\end{eqnarray}

\noindent{}but this is often infeasible since the computation and storage of the Hessian $\H_t$ is intractable for large $n$. The usual way to deal with this problem by using limited-memory quasi-Newton methods such as L-BFGS \citep{liu1989limited} is not currently in favor in deep learning, not the least due to (i) the stochasticity of $\nabla f_t(\vc{x}_t)$, (ii) ill-conditioning of $f$ and (iii) the presence of saddle points as a result of the hierarchical geometric structure of the parameter space \citep{fukumizu2000local}. 
Despite some recent progress in understanding and addressing the latter problems \citep{bordes2009sgd,dauphin2014identifying,choromanska2014loss,dauphin2015rmsprop}, 
state-of-the-art optimization techniques tend to approximate the inverse Hessian in a reduced way, e.g., by considering only its diagonal to achieve adaptive learning rates.  
 
In this paper, we argue that not only the update of $\vc{x}_t$ given $\nabla f_t(\vc{x}_t)$ is crucial, but the selection of the batch $\left\{ \psi^b_{i=1} \right\} \sim \D^b $ used to compute $\nabla f_t(\vc{x}_t)$ also greatly contributes to the overall performance. 

\section{AdaDelta and Adam}

We start by describing the Adadelta and Adam algorithms, which we both build upon and use as the baseline for our experiments. 
Other related work (most of which appeared as parallel ICLR submissions) is described in Section 7. 

\begin{algorithm}[tb!]
\caption{The original AdaDelta algorithm}
\label{algo_adadelta}
\begin{algorithmic}[1]
\STATE{\textbf{given} $\rho = 0.95, \epsilon = 10^{-6}$} \label{adadelta-Given}
\STATE{\textbf{initialize} time step $t \leftarrow 0$, parameter vector $\vc{x}_{t=0} \in \R^n$, first moment vector $\vc{s}_{t=0} \leftarrow \vc{0}$, second moment vector $\vc{v}_{t=0} \leftarrow \vc{0}$}
\REPEAT
	\STATE{$t \leftarrow t + 1$}
	\STATE{$\nabla f_t(\vc{x}_{t-1}) \leftarrow  SelectBatch(\vc{x}_{t-1})$} \COMMENT{select batch and return the corresponding gradient} \label{callsel}
	\STATE{$\vc{g}_t \leftarrow \nabla f_t(\vc{x}_{t-1})$}
	\STATE{$\vc{v}_t \leftarrow \rho \vc{v}_{t-1} + (1 - \rho) \vc{g}^2_t $} \COMMENT{here and below all operations are element-wise} \label{adadelta-vvector} 
	\STATE{$\Delta \vc{x}_t \leftarrow - \frac{\sqrt{\vc{s}_{t-1} + \epsilon}}{\sqrt{\vc{v}_t + \epsilon}}\vc{g}_t $}  \label{adadelta-dvector}
	\STATE{$\vc{s}_t \leftarrow \rho \vc{s}_{t-1} + (1 - \rho) \Delta \vc{x}^2_t $}  \label{adadelta-st}
	\STATE{$\vc{x}_t \leftarrow \vc{x}_{t-1} + \Delta \vc{x}_t$}
\UNTIL{ \textit{stopping criterion is met} }
\STATE{\textbf{return} \,optimized parameters $\vc{x}_t$}
\end{algorithmic}
\end{algorithm}

\begin{algorithm}[tb!]
\caption{The original Adam algorithm}
\label{algo_adam}
\begin{algorithmic}[1]
\STATE{\textbf{given} $\alpha = 0.001, \beta_1 = 0.9, \beta_2 =0.999, \epsilon = 10^{-8}$} \label{adam-Given}
\STATE{\textbf{initialize} time step $t \leftarrow 0$, parameter vector $\vc{x}_{t=0} \in \R^n$,  first moment vector $\vc{m}_{t=0} \leftarrow \vc{0}$, second moment vector  $\vc{v}_{t=0} \leftarrow \vc{0}$}
\REPEAT
	\STATE{$t \leftarrow t + 1$}
	\STATE{$\nabla f_t(\vc{x}_{t-1}) \leftarrow  SelectBatch(\vc{x}_{t-1})$}  \COMMENT{select batch and return the corresponding gradient}
	\STATE{$\vc{g}_t \leftarrow \nabla f_t(\vc{x}_{t-1})$}
	\STATE{$\vc{m}_t \leftarrow \beta_1 \vc{m}_{t-1} + (1 - \beta_1) \vc{g}_t $} \label{adam-mom1} \COMMENT{here and below all operations are element-wise}
	\STATE{$\vc{v}_t \leftarrow \beta_2 \vc{v}_{t-1} + (1 - \beta_2) \vc{g}^2_t $} \label{adam-mom2}
	\STATE{$\hat{\vc{m}}_t \leftarrow \vc{m}_t/(1 - \beta_1^t) $} \COMMENT{here, $\beta_1$ is taken to the power of $t$} \label{adam-corr1}
	\STATE{$\hat{\vc{{v}}}_t \leftarrow \vc{v}_t/(1 - \beta_2^t) $} \COMMENT{here, $\beta_2$ is taken to the power of $t$} \label{adam-corr2}
	\STATE{$\vc{x}_t \leftarrow \vc{x}_{t-1} - \alpha \hat{\vc{m}}_t / (\sqrt{\hat{\vc{v}}_t} + \epsilon)$}
\UNTIL{ \textit{stopping criterion is met} }
\STATE{\textbf{return} \,optimized parameters $\vc{x}_t$}
\end{algorithmic}
\end{algorithm}

The AdaDelta algorithm \citep{zeiler2012adadelta} is based on AdaGrad \citep{duchi2011adaptive}, 
where the learning rate $\eta_t$ is set to be $\eta/\sqrt{\sum_{i=1}^t({\nabla f_i(\vc{x}_i)})^2}$ 
with the purpose to estimate the diagonal of the Hessian of the functions $f_i$ \citep{duchi2011adaptive}. 
This idea can be traced back to variable metric methods, such as the BFGS family of algorithms \citep{fletcher1970new}. The accumulation of gradient amplitudes from $t=1$ can negatively impact the search if the old information is irrelevant for the current objective function landscape. 
Similarly to the approach proposed by \citet{schaul2012no}, AdaDelta attempts to overcome this issue by decaying the information about gradient amplitudes (stored in vector $\vc{v}_t$, see line \ref{adadelta-vvector} in Algorithm \ref{algo_adadelta}) with a factor of $1-\rho$ defining the time window. Additionally, AdaDelta stores the information about parameter changes $\Delta \vc{x}_t$ (stored in vector $\vc{s}_t$, see line \ref{adadelta-st}) to recompute the new $\Delta \vc{x}_t$ as $- \frac{\sqrt{\vc{s}_{t-1} + \epsilon}}{\sqrt{\vc{v}_t + \epsilon}}\vc{g}_t $. 
The purpose of this normalization is to capture the change of gradient amplitudes and to adjust the learning rates accordingly. 

The Adam algorithm \citep{kingma2014adam} considers two momentum vectors, 
the first momentum $\vc{m}_t$ for gradients with a decay factor 1 - $\beta_1$ (see line \ref{adam-mom1} 
in Algorithm \ref{algo_adam}) and the second momentum $\vc{v}_t$ for gradient amplitudes 
with a decay factor 1 - $\beta_2$ (see line \ref{adam-mom2}). 
In order to overcome the initialization bias introduced by setting $\vc{m}_{t=0}$ and $\vc{v}_{t=0}$ to zero vectors, the authors proposed to use bias-corrected estimates $\hat{\vc{m}}_t$ and $\hat{\vc{v}}_t$ (see lines \ref{adam-corr1}-\ref{adam-corr2}). The change of $\vc{x}_t$ is performed according to the first bias-corrected momentum $\hat{\vc{m}}_t$, with the learning rates set according to the second bias-corrected momentum $\hat{\vc{v}}_t$. 

The principal reason for selecting AdaDelta and Adam for our experiments is 
their claimed robustness which should reduce the problem of 
hyperparameter tuning to obtain satisfactory
performance. 

\begin{algorithm}[tb!]
\caption{Batch Selection Procedure SelectBatch()}
\label{batchsel}
\begin{algorithmic}[1]
\STATE{\textbf{given} parameter vector $\vc{x} \in \R^n$} 
\STATE{\textbf{global parameters} number of datapoints $N$, datapoints $d_i$ for $i=1,\ldots,N$; batch  size $b$; counters $c$, $c_{e}$, $c_{s}$, $c_{r}$; two-dimensional array $l$ with $l^{(1)}_{(1:N)}$ storing loss values and $l^{(2)}_{(1:N)}$ storing datapoint indexes; selection pressure $s_e$; 
selection probabilities $p_i$ and cumulative selection probabilities $a_i$ defined for $i$-th loss ranked datapoints; period of sorting $T_s$;  frequency of loss recomputation $r_{freq}$ for $r_{ratio}\cdot N$ datapoints with the greatest known loss values}
\STATE{\textbf{first call initialization} $c \leftarrow 0 $, $c_{e} \leftarrow - N$, $c_{s} \leftarrow 0 $, $c_{r} \leftarrow 0$, 
$l^{(1)}_{(1:N)}=\infty$, $l^{(2)}_{(1:N)}=1:N$
}
	\STATE{$c \leftarrow c + b$}
	\IF{$c - c_{e} > N $}
		\STATE{$c_{e} \leftarrow {c}, e \leftarrow c_{e}$ div $N$} \COMMENT{once per epoch, we recompute selection probabilities} 	
		\STATE{use Eq. (5) to compute $s_e$}	\COMMENT{$O(1)$}	\label{computee}
		\STATE{use Eq. (4) to compute $p_i, i=1,\ldots,N$ }	\COMMENT{$O(N)$} \label{computel}
		\STATE{$a_{j} \leftarrow \sum_{i=1}^j p_i$ for $j=1,\ldots,N$} \COMMENT{$O(N)$} \label{computec}
	\ENDIF{}
	\IF{$c - c_{s} > T_s$}	
		\STATE{$c_{s} \leftarrow {c}$} \COMMENT{every $T_s$ evaluated datapoints, we sort datapoints w.r.t. the latest known loss}
		\STATE{$l \leftarrow sort(l)$ in descending order w.r.t. loss values stored in array $l^{(1)}_{(:)}$}	\COMMENT{$O(N\log(N))$} \label{sort1}
	\ENDIF{}
	\IF{$c - c_{r} > N/r_{freq}$}	
		\STATE{$c_{r} \leftarrow {c}$} \COMMENT{$r_{freq}$ times per epoch, we recompute loss values for the top $r_{ratio} \cdot N$ datapoints}
		\FOR{i = 1 to $r_{ratio} \cdot N$} 
				\STATE{$l^{(1)}_{i} \leftarrow \psi_i (\vc{x})$ w.r.t. datapoint with index $l^{(2)}_{i}$} 
				\COMMENT{we do it in biggest possible batches} \label{recomputeloss}
		\ENDFOR
		\STATE{$l \leftarrow sort(l)$ in descending order w.r.t. loss values stored in array $l^{(1)}_{(:)}$}	\COMMENT{$O(N\log(N))$} \label{sort2}
	\ENDIF{}
	\FOR{i = 1 to $b$} 
	\IF{Default random selection}
		\STATE{sample datapoint index $d_i \in [1, N]$ uniformly at random}
	\ENDIF{}
	\IF{Default shuffle selection}
		\STATE{sample datapoint index $d_i \in [1, N]$ uniformly at random  without repetitions in one epoch}
	\ENDIF{}
	\IF{Loss-dependent selection}
			\STATE{sample random $r \in [0,1)$, find the lowest rank index $i_{sel}$ s.t. $r < a_{i_{sel}}$} \COMMENT{$O(\log(N))$} \label{computeisel}
			\STATE{$d_i \leftarrow l^{(2)}_{i_{sel}}$} \label{computediisel}
		\ENDIF{}
		\STATE{associate $\psi_i (\vc{x})$ with the loss function computed on the datapoint with index $d_i$}
	\ENDFOR
\STATE{ $f \leftarrow \frac{1}{b}\sum_{i=1}^b \psi_i (\vc{x})$}	\COMMENT{forward pass}
\FOR{i = 1 to $b$} 
		\STATE{$l^{(1)}_{i_{sel}} \leftarrow \psi_i (\vc{x})$}	\COMMENT{latest known loss}
\ENDFOR
\STATE{\textbf{return} \,$\nabla f(\vc{x})$}	\COMMENT{compute and return gradient}

\end{algorithmic}
\end{algorithm}

\section{Online Batch Selection}

All approaches mentioned in the previous section attempt to deal with stochasticity arising from the fact that the noise-less $\nabla f(\vc{x}_t)$ is often intractable to compute. Instead, $\nabla f_t(\vc{x}_t)$ is computed w.r.t. a batch $\left\{ \psi^b_{i=1} \right\} \sim \D^b $ (see line 5 in Algorithms 1 and 2). It is common to reshuffle the whole dataset once per epoch and divide it into a number of batches with size $b$ and thus to  schedule the appearance of datapoints. As different datapoints correspond to different loss function values (e.g., some are more difficult to classify), gradient amplitudes and directions may also appear to be significantly different. This motivated the use of momentum both for gradient directions and amplitudes to smooth the incoming information. 

We propose to online select batches (both size and datapoints) for an algorithm $\A$  to maximize its progress (e.g., defined by the objective function $f$ or cross-validation error) over the resource budget (e.g., the number of evaluated datapoints or time).  
 More formally, this problem can be formulated in the framework of reinforcement learning \citep{sutton1998reinforcement}. 
In this paper, we provide an initial study of this problem in the context of deep learning to explore possible design choices starting from the simplest ones. Related works on batch selection are discussed in section \ref{relwork}.

Let us assume for simplicity that the number of available training datapoints is bounded by $N$.
At each time step $t$, the probability of the $i$-th datapoint to be selected to compose a batch of size $b$ is denoted by $p_{i}$. By setting $p_{i}$ to 0 for all already selected datapoints and to $\frac{1}{N+1-t}$ otherwise, the whole set will be enumerated in $N/b$ steps (denoted here as one epoch). Clearly, if we set the selection probability $p_{i}$ of some datapoints to 0, then, we reduce the size of the training set by eliminating these datapoints from the batch selection. A smaller training set is typically easier to fit but also to over-fit. A possible strategy is to first select the datapoints with the greatest contributions to the objective function; e.g., one could always select the datapoint with the greatest loss. 
However, after the exploitation of that datapoint is performed and its loss is (likely) reduced, the datapoint with the next greatest loss would need to be identified and any datapoint from the training set is a potential candidate. While one can rely on the assumption that ranks of losses over datapoints change slowly in time, the exploitation-exploration dilemma is inevitable.

The procedure we propose for batch selection is given in Algorithm \ref{batchsel}, 
and we can integrate it into AdaDelta and Adam by calling it in line \ref{callsel} of 
Algorithms \ref{algo_adadelta} and \ref{algo_adam}. When datapoints are shuffled or selected uniformly at random (i.e., when the default AdaDelta and Adam are used), the first 17 lines of Algorithm \ref{batchsel} should be skipped. 
  
We propose a simple rank-based batch selection strategy, where all training datapoints $k$ are ranked (sorted) in descending order w.r.t. their latest computed $\psi_k(\vc{x})$ (see lines \ref{sort1} and \ref{sort2} in Algorithm \ref{batchsel}). 
Each $i$-th ranked datapoint is selected with probability $p_i$. 
We propose to exponentially decay $p_i$ with its rank $i$ such that ${p_N}=\frac{p_1}{s_e}$, 
where ${s_e}$ is the ratio between the greatest and smallest probabilities of selection at epoch $e$. 
Thus, between any $i$ and $i+1$, the probability drops by a factor of $\exp(\log(s_e)/N)$. 
When normalized to sum up to 1.0, the probability of $i$-th ranked datapoint to be 
selected is defined as 

\begin{eqnarray}
	\label{eq:prob}
	p_i = \frac{1 / \exp(\log(s_e)/N)^i}{\sum_{i=1}^{N} 1 / \exp(\log(s_e)/N)^i},
\end{eqnarray}

Intuitively, this selection pressure parameter $s_e$ affects the effective size of the training data we use. Let $N_{eff}$ denote the number of unique datapoints selected among the last $N$ selections; then, $N_{eff}$ is likely (but not necessarily, see section \ref{beffb} in the supplementary material) to decrease when $s$ increases.  

To imitate the increase of $N_{eff}$ at each epoch $e$, from epoch ${e_{0}}$ to epoch ${e_{end}}$, we exponentially decrease the selection pressure from $s_{e_0}$ to $s_{e_{end}}$ with 

\begin{eqnarray}
	\label{eq:ev1}
	s_e=s_{e_0} \exp(\log(s_{e_{end}}/s_{e_0})/({e_{end}}-{e_0}))^{{e_{end}}-{e_0}},
\end{eqnarray}

and then substitute it into (4) to compute it once per epoch. 
The realization of datapoint selection is quite straightforward: given an array $a_{j} \leftarrow \sum_{i=1}^j p_i$ for $j=1,\ldots,N$  (once per epoch when $s_e$ is updated, see lines \ref{computee}-\ref{computec}), we uniformly at random generate $r \in [0,1)$ and find the lowest index $i_{sel}$ such that $r < a_{i_{sel}}$ (see line \ref{computeisel}). The datapoint to be selected is the one whose rank w.r.t. the latest known loss value is $i_{sel}$ (see line \ref{computediisel}). 

The latest loss computed for a datapoint is only an estimate of its current loss.  
In order to improve the quality of the ranking, we propose to periodically, $r_{freq}$ times per epoch, recompute the loss for $r_{ratio} \cdot N$ datapoints with the greatest latest known loss (see lines \ref{recomputeloss}-\ref{sort2}). 

\section{Time Complexity}
The computational complexity of datapoint selection scales as $O(\log(N))$ per datapoint using a bisection-based search in the array $a$ whose values are increasing with rank. The sorting procedure scales as $O(N\log(N))$, but is performed periodically every $T_s$ evaluated datapoints (which in turn reduces the constant factor because the array is partially sorted). The time overhead of the selection procedure without recomputing loss values (see lines \ref{recomputeloss}-\ref{sort2}) is roughly 5\% when measured on a GPU for MNIST. If the cost of computing training and validation losses (for plotting) is included, then the relative cost is smaller than 5\%. 

The procedure of recomputing loss values (see lines \ref{recomputeloss}-\ref{sort2}) is optional. 
It is parameterized by its frequency $r_{freq}$ and the amount of datapoints $r_{ratio} \cdot N$; it costs $r_{freq} \cdot r_{ratio}$ forward passes per datapoint. In practice, if the maximum feasible  
batch size is used and the recomputation is performed every two epochs ($r_{freq}=0.5$) for the whole dataset ($r_{ratio}=1.0$), 
then the total overhead is expected to be roughly $20\%$ (i.e., by a factor of 1.2 more expensive) w.r.t. the default algorithms.  If the cost of computing training and validation losses (for plotting) is included (e.g., it is excluded in Figure \ref{Figure1}), then the relative cost is smaller than $20\%$. 

\section{Experimental Results}

\begin{figure*}[p]
\begin{center}
\includegraphics[width=0.495\textwidth]{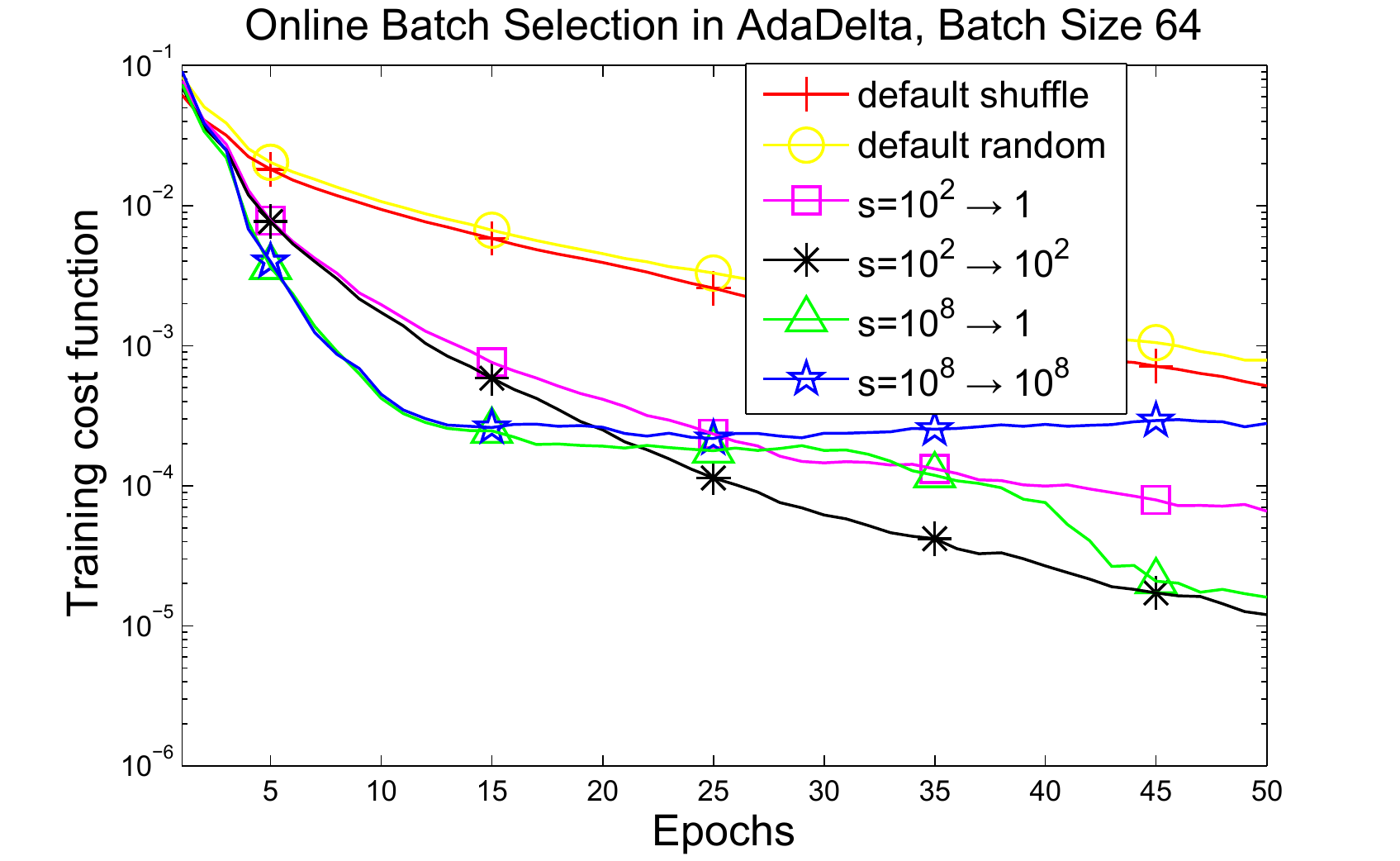}
\includegraphics[width=0.495\textwidth]{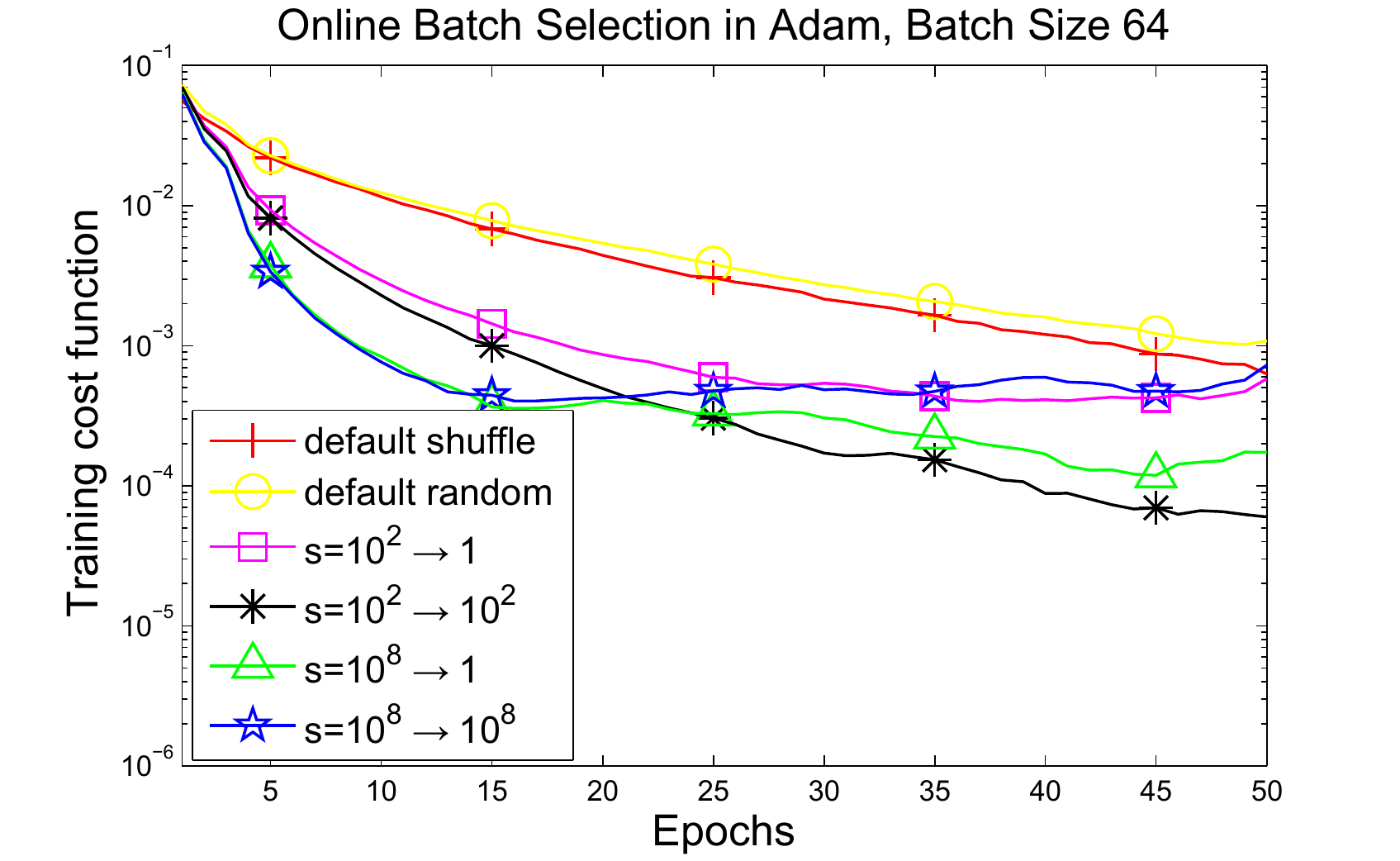}\\
\includegraphics[width=0.495\textwidth]{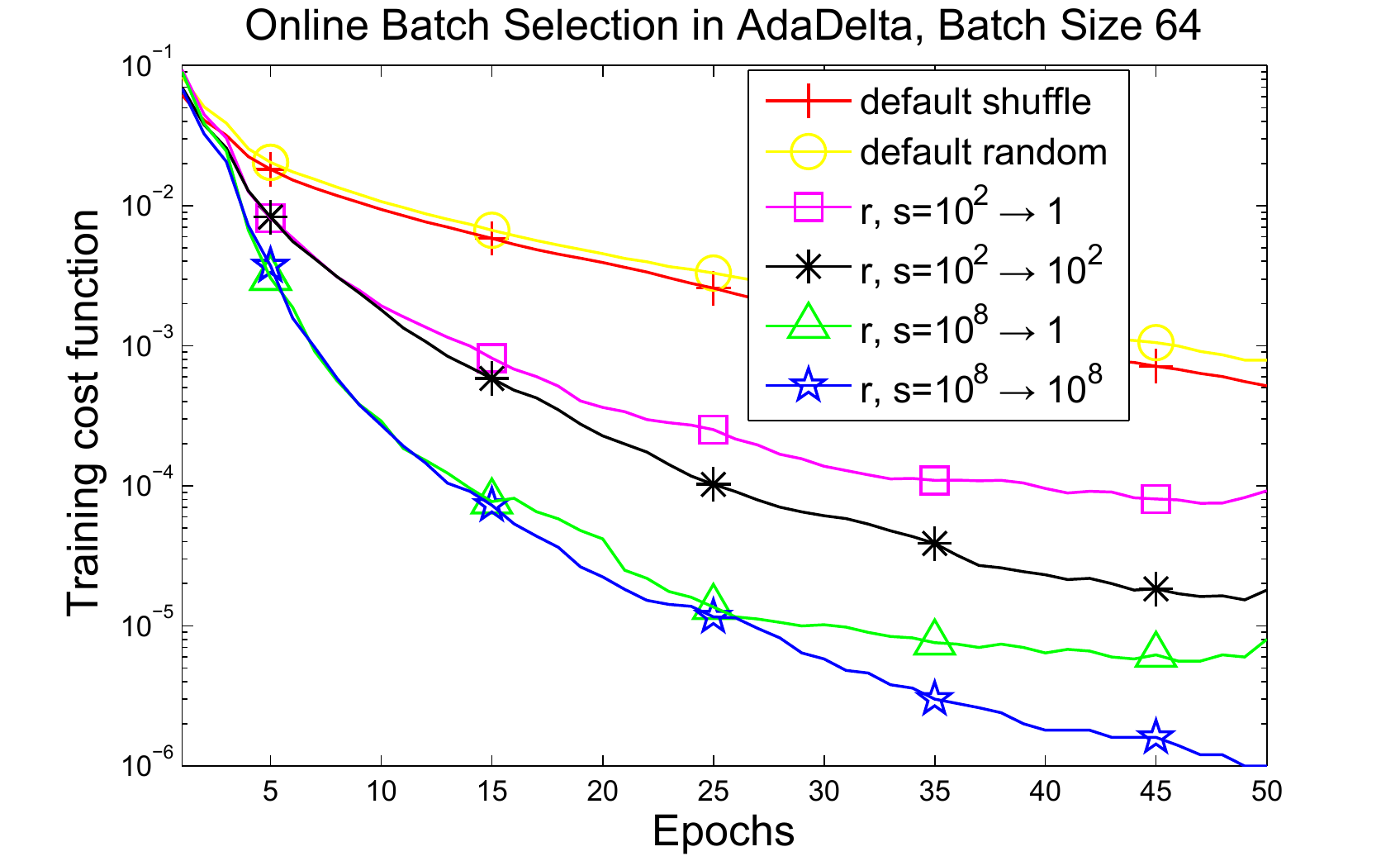}
\includegraphics[width=0.495\textwidth]{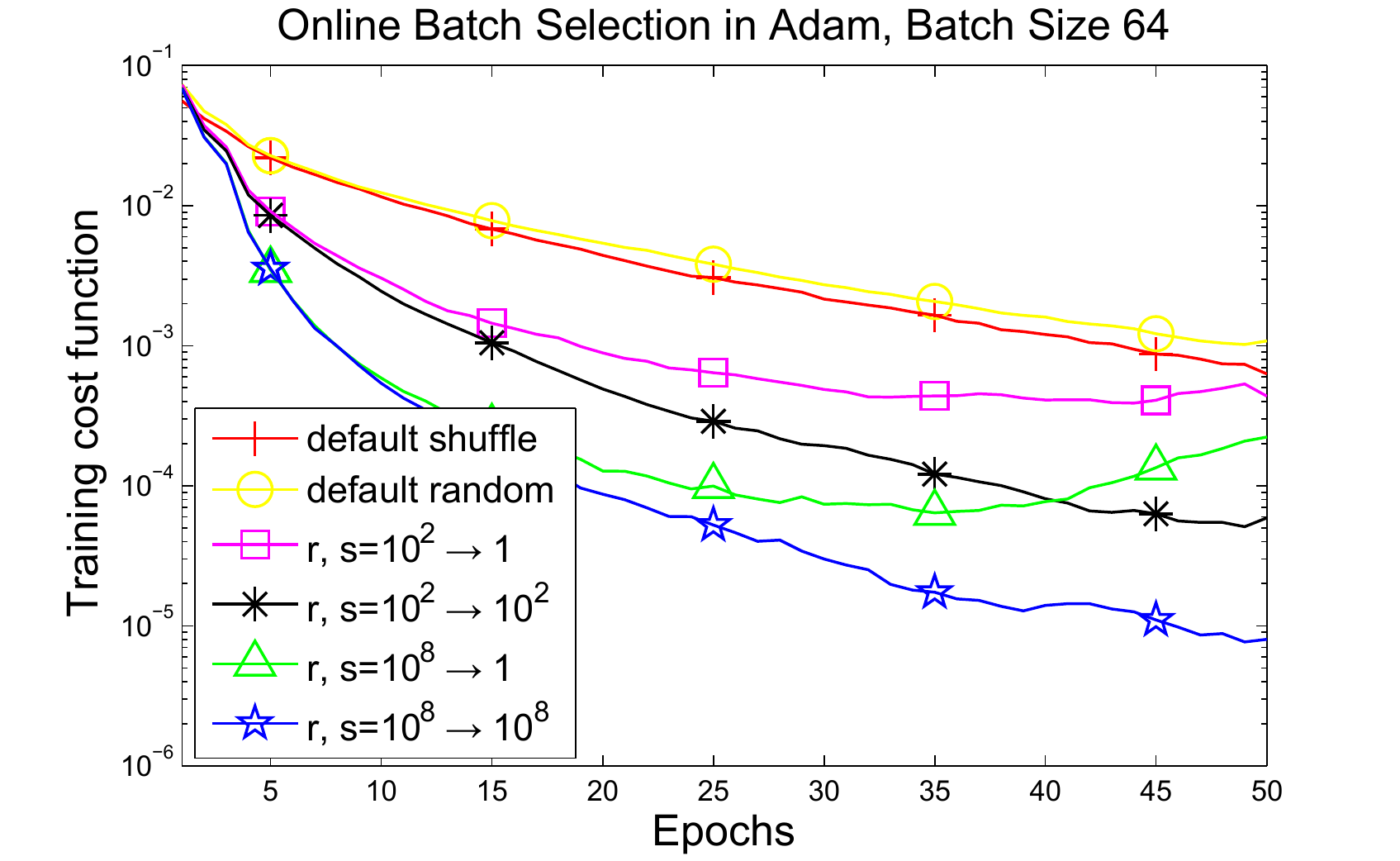}\\
\includegraphics[width=0.495\textwidth]{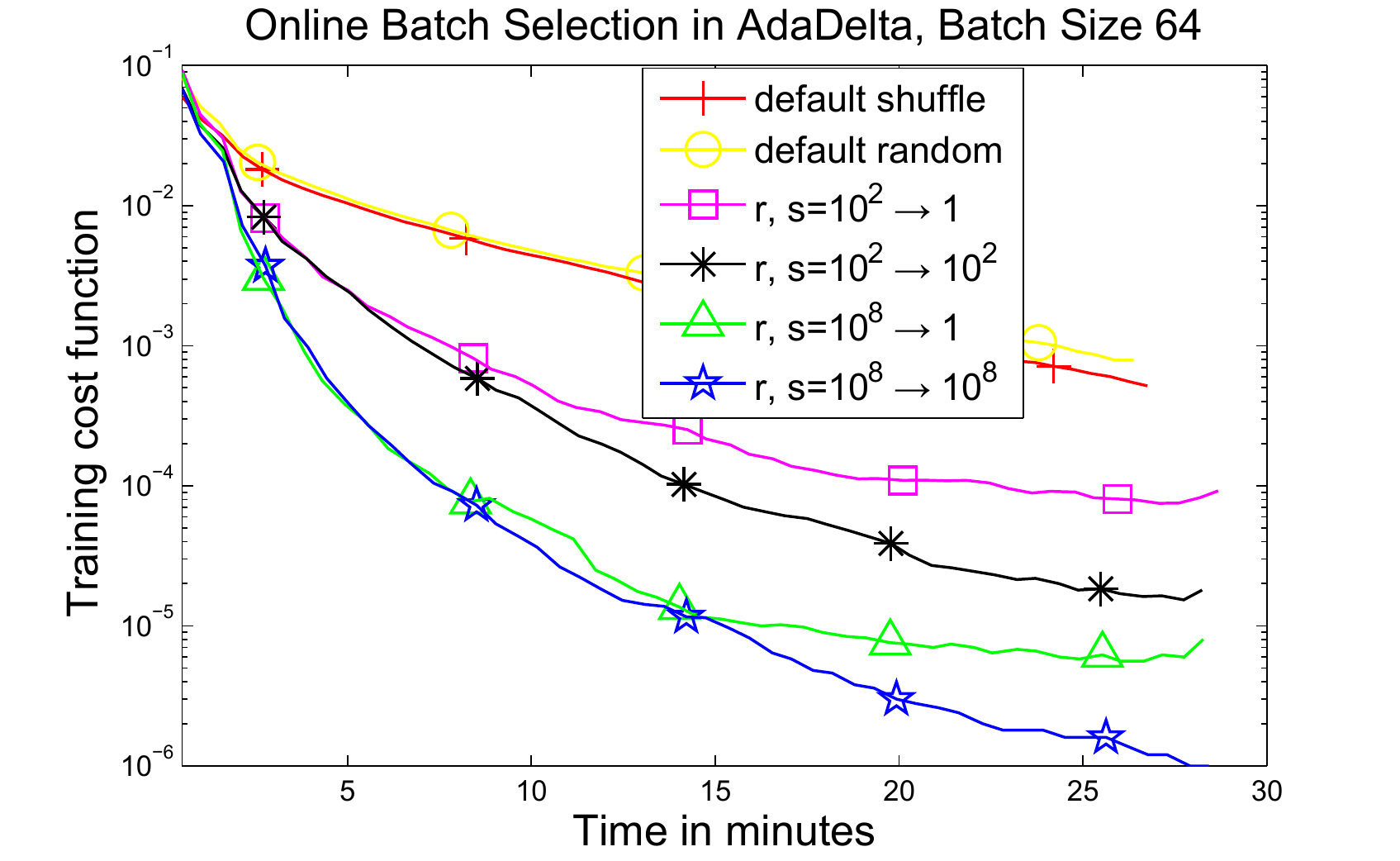}
\includegraphics[width=0.495\textwidth]{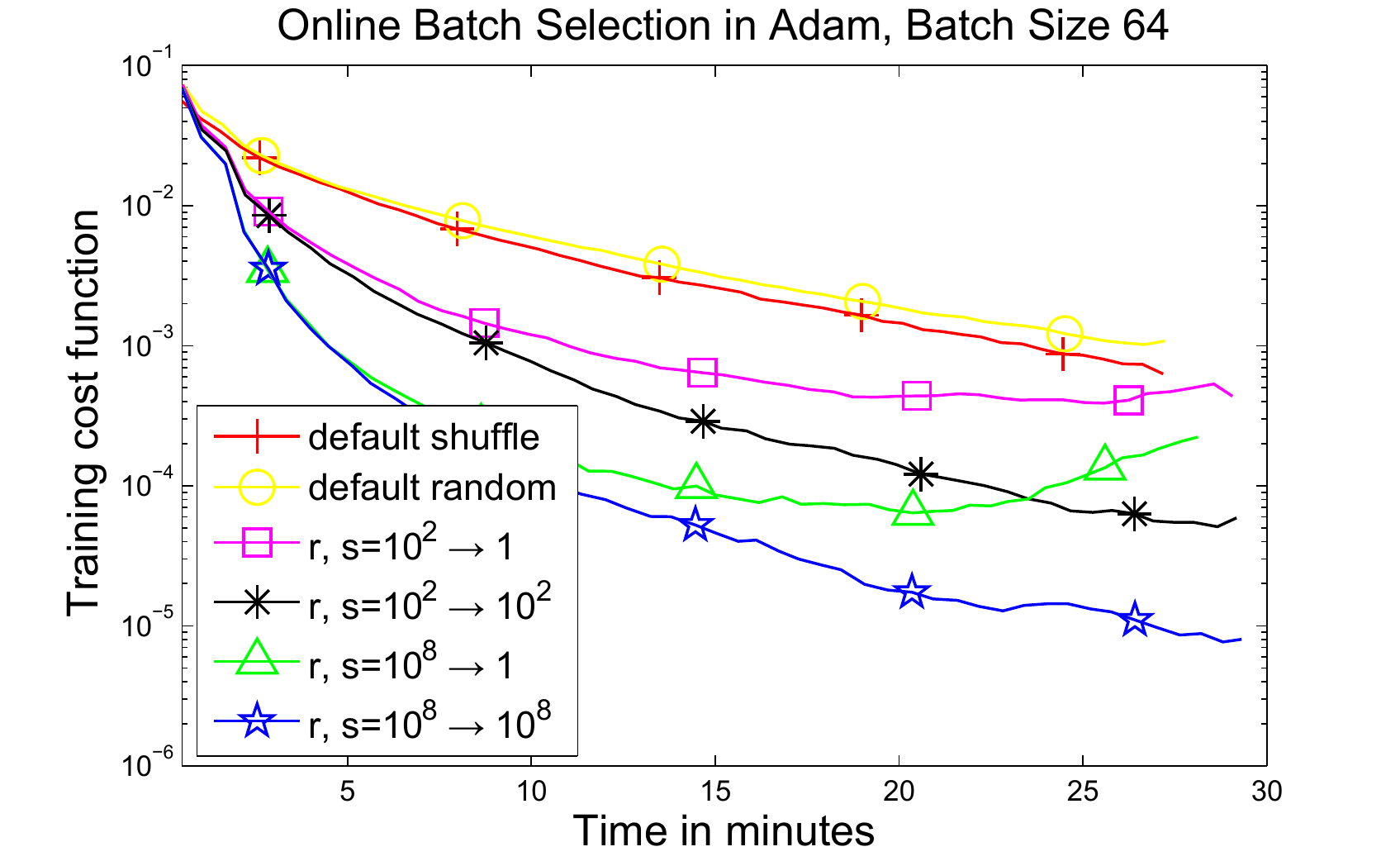}\\
\includegraphics[width=0.495\textwidth]{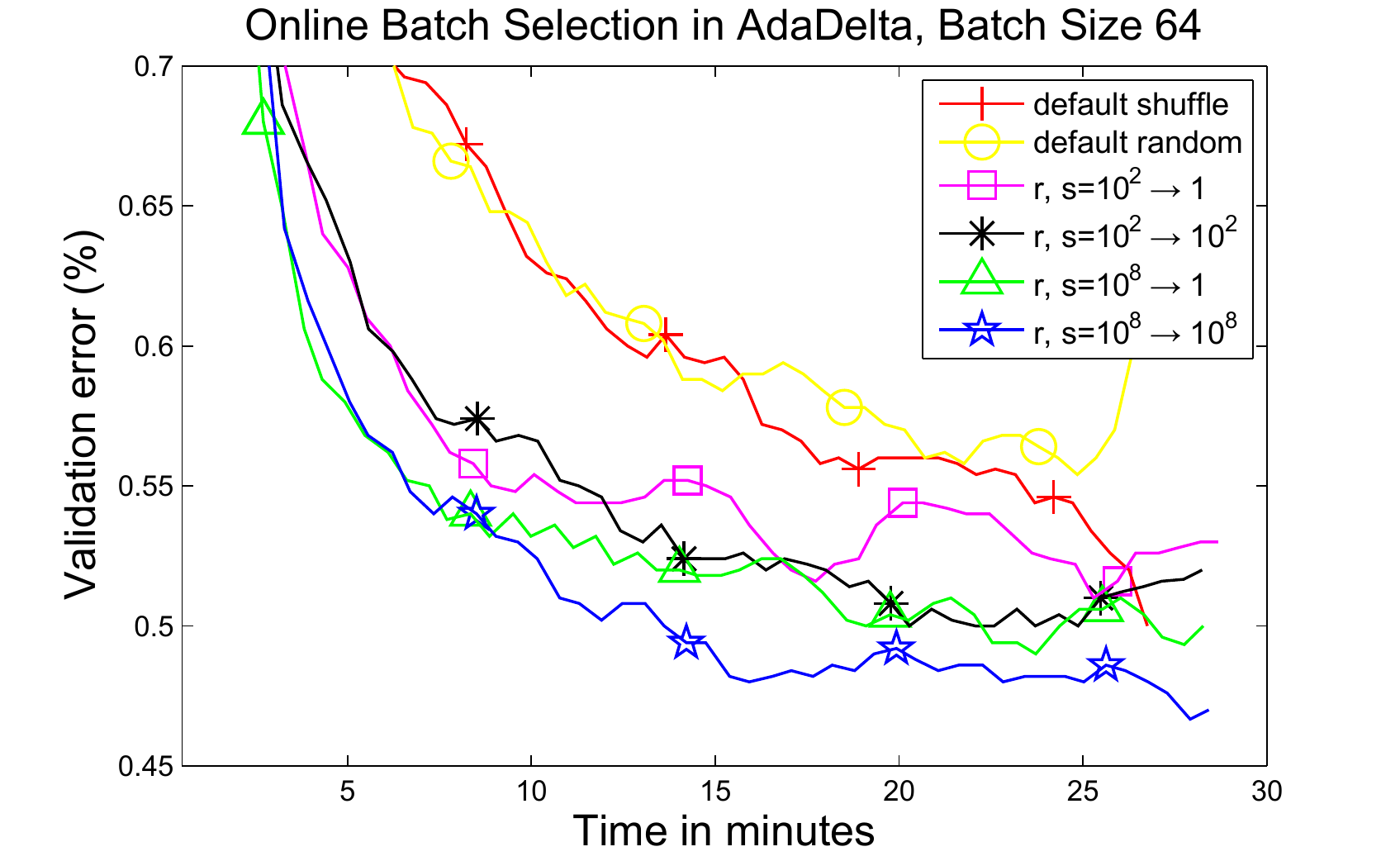}
\includegraphics[width=0.495\textwidth]{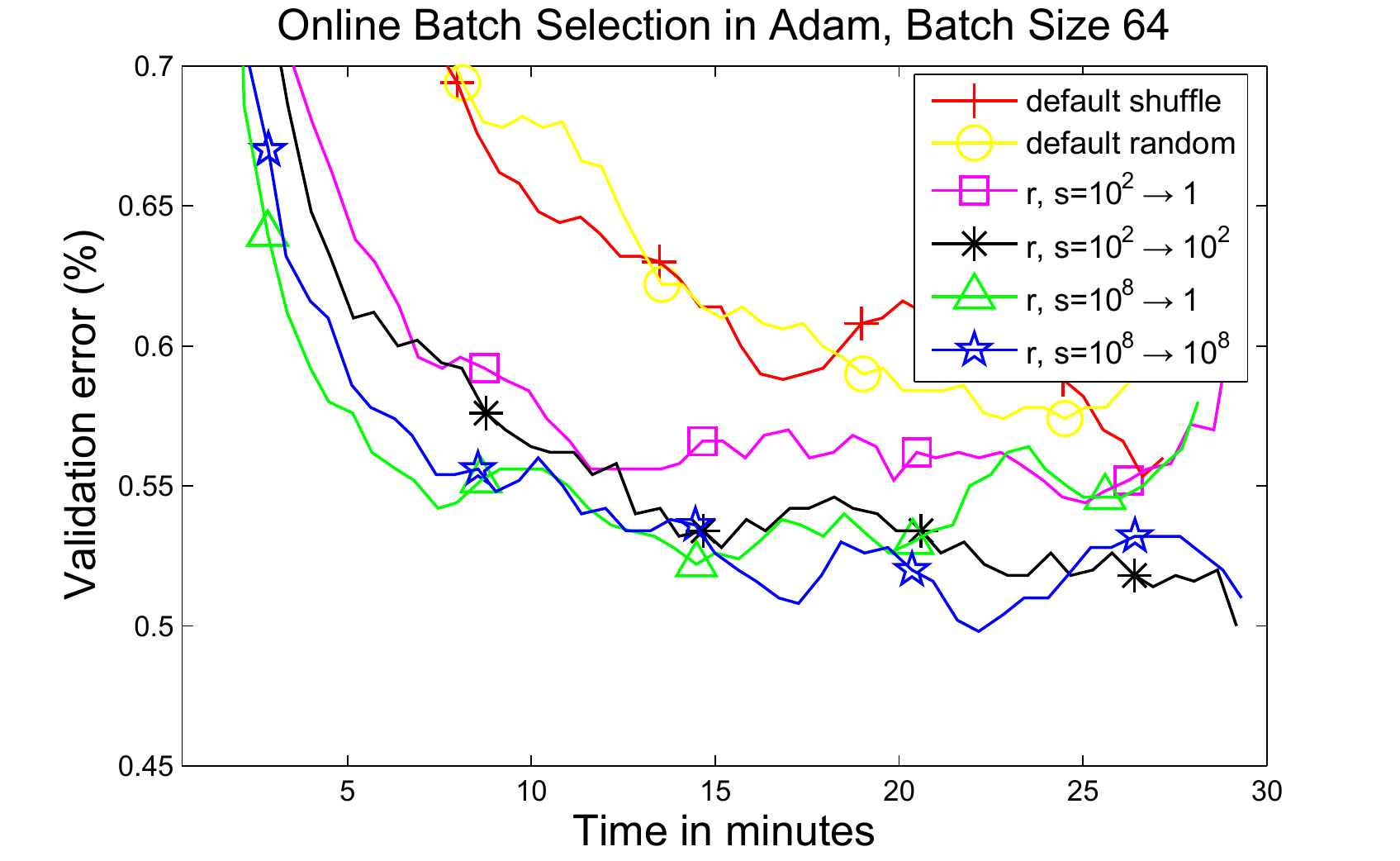}\\
\end{center}
\caption{
Convergence curves of AdaDelta (\textbf{Left Column}) and Adam (\textbf{Right Column}) on MNIST dataset. 
The original algorithms are denoted by ``default shuffle'' (respectively, ``default with random'') when datapoints are shuffled and selected sequentially (respectively, uniformly at random). 
The value of $s$ denotes the ratio of probabilities to select the  training datapoint with the greatest latest known loss rather than the smallest latest known loss. 
Legends with 
$s=s_{e_0} \rightarrow s_{e_{end}}$ correspond to an exponential change of $s$ from $s_{e_0}$ to $s_{e_{end}}$ as a function of epoch index $e$, see Eq. (\ref{eq:ev1}). 
Legends with the prefix ``r'' correspond to the case when the loss values for $r_{ratio} \cdot N=1.0N$ datapoints with the greatest latest known loss 
 are recomputed $r_{freq}=0.5$ times per epoch. 
All curves are computed on the whole training set and smoothed by a moving average of the median of 11 runs.
}
\label{Figure1}
\end{figure*}

\begin{figure*}[t]
\begin{center}
\includegraphics[width=0.65\textwidth]{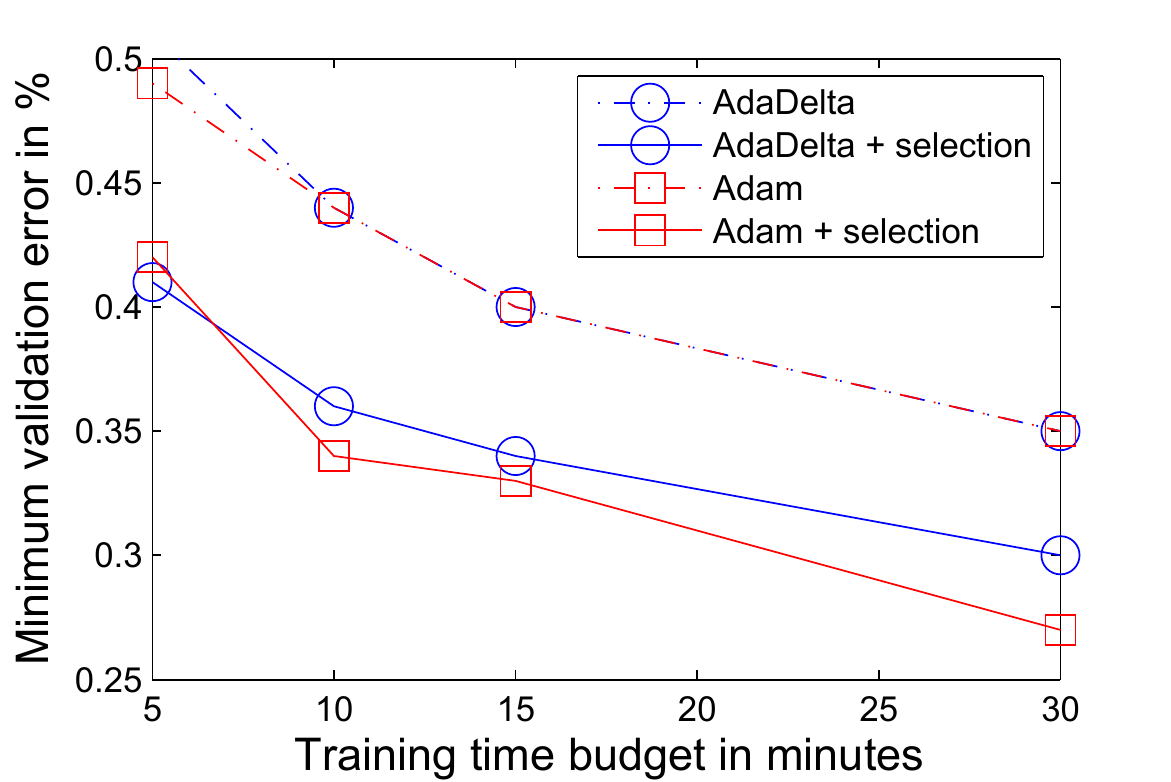}
\end{center}
\caption{
Minimum validation error found on MNIST for deep neural network trained with a given budget of time. 
The results correspond to the best hyperparameter settings found by CMA-ES after 1000 function evaluations.  
The original Adam and Adadelta algorithms are denoted by dotted lines, their versions with batch selection are denoted by 
solid lines. 
}
\label{FigureMnistAll}
\end{figure*}

We empirically investigate the impact of the proposed online batch selection strategies 
when combined with AdaDelta and Adam. 
We conducted our experiments on the standard MNIST dataset of handwritten digits \citep{lecun1998gradient} 
with 50.000 training, 10.000 validation and 10.000 test datapoints. 
We do not employ any data augmentation and pre-processing procedures.  
All experiments were performed using the python library Lasagne\footnote{\url{https://github.com/Lasagne/Lasagne}}; we provide the source code to reproduce our experiments online at \url{https://sites.google.com/site/batchsel/} (anonymous for the reviewers). 
In all experiments, we use Batch Normalization \citep{ioffe2015batch} as implemented by \cite{courbariaux2015binaryconnect} in Lasagne because i) we found that it consistently improves the error rate by roughly 0.1-0.15\%, 
and ii) to demonstrate the compatibility of our approach with the existing acceleration techniques. 

We performed our experiments with various values of the batch size $b$, but mainly report the results for $b=64$ 
(see Figure \ref{Figure1}) because they demonstrate better generalization errors over time. The results for $b=16$ and $b=256$ are given in the supplementary material (Figures \ref{Figure2} and \ref{Figure3}, respectively). 

Figure \ref{Figure1} shows that a 5-fold speedup can be achieved 
i) during the first 10 epochs if  
the recomputation procedure is not considered (see our early experiments in Figure \ref{Figure1} (top row)), 
and, 
ii) during the whole run if the recomputation procedure is considered 
(see the remaining rows in Figure \ref{Figure1}). 
Figure \ref{Figure1} (top row) is reported mainly to support the motivation for the recomputation procedure.  

Figure \ref{Figure1} (top row) shows the cost function (cross-entropy loss) over epochs for the original AdaDelta and Adam 
(when datapoints are shuffled and selected sequentially or uniformly at random) and 
their accelerated versions with our batch selection procedure. 
Our original approach was to have a fixed $s$ 
(see, e.g., $s=10^2 \rightarrow 10^2$ and $s=10^8 \rightarrow 10^8$ in the figure) and we found that $s=10^2$ 
significantly speeds up both AdaDelta and Adam. 
Greater values, such as $s=10^8$, lead to even faster convergence during the first 10 epochs. 
However, the cost function stops decaying at later epochs and a divergence may occur (more likely for greater values of $s$).  
This effect might partially explain why previous attempts (if any) to perform batch selection 
were not successful: the use of very strong selection might lead to divergence.  
To reduce this problem, we introduced a decreasing $s$ over epochs as described in Eq. (\ref{eq:ev1}). 
The exponential decrease of $s$ from $10^2$ to $1$ over 50 epochs (denoted as $s=10^2 \rightarrow 1$) 
corresponds to a standard uniform selection at the last epoch such that most of the training datapoints will be considered. 
Using  $s=10^2 \rightarrow 1$ corresponds to a weaker selection 
and a reduced speedup compared to $s=10^2 \rightarrow 10^2$. 
However, $s=10^8 \rightarrow 1$ combines the fast convergence at the first 10 epochs with a steady improvement at the later epochs. 

To further improve the convergence for greater values of $s$, 
we proposed to additionally, $r_{freq}=0.5$ times per epoch, recompute the loss for $r_{ratio} \cdot N=1.0N$ datapoints with the greatest latest known loss. 
Figure \ref{Figure1} (second row) shows the results when this recomputation (denoted by a prefix ``r'')  is applied. It does not change the results much for $s=10^2 \rightarrow 10^2$ and $s=10^2 \rightarrow 1$, but drastically improves the results for $s=10^8 \rightarrow 10^8$. 
The reason for this improvement is a more consistent ranking of the datapoints.

Any improvement in terms of the number of epochs to achieve certain training loss values is interesting for algorithm design. However, the performance over time is often of more practical interest, and we show it in Figure \ref{Figure1} (third row). 
This figure demonstrates that the relative overhead coming from the batch selection procedure 
is rather negligible and is greatly compensated by the overall speedup of a factor of 5 observed 
after 5 epochs for both AdaDelta and Adam.

In order to check whether the faster convergence leads to an overfitting, 
we present Figure \ref{Figure1} (last row) with the error measured on the validation dataset over time. The results suggest that the accelerated AdaDelta and Adam achieve better validation errors faster, by a factor up to 5.  

It could be possible that the obtained results are just artifacts of the used hyperparameters settings. 
To check this hypothesis, we performed hyperparameter optimization of both networks and learning strategies to minimize  
the validation error of both the original Adam and Adadelta and their variants with batch selection. 
Figure \ref{FigureMnistAll} confirms that our proposed batch selection is beneficial for training networks on MNIST and leads 
to a speedup of peak performance by a factor of 3. The details of the employed hyperparameter optimization algorithm (Covariance Matrix Adaptation Evolution Strategy, CMA-ES, by \cite{hansen2001completely}), as well as all experimental settings, are given in the supplementary material (Section \ref{sec:hyperparam_opt}).  

\section{Related Work}\label{relwork}

The use of uniform sampling of training examples or dual coordinates is a common way to deal with large-scale stochastic optimization problems. 
Although the sampled stochastic quantity is an unbiased estimate of the corresponding true quantity, the variance of the resulting stochastic quantity may be very high since individual estimates can vary significantly over different examples \citep{zhao2014stochastic}. 
The idea to use a non-uniform sampling is known in derivative-free optimization (more specifically, in evolutionary computation) as selection \citep{back1996evolutionary} and is used in state-of-the-art optimizers to bias the search towards better solutions
\citep{hansen2001completely}. 
Recently, the idea to bias the selection of more important training examples or dual coordinates has 
gained popularity in coordinate-descent methods \citep{nesterov2012efficiency} and stochastic gradient methods \citep{needell2014stochastic,zhao2014stochastic,schmidt2015non}. 
These selection strategies are often described in the framework of importance sampling~\citep{zhao2014stochastic}. They often consist of i) 
biasing the sampling towards the Lipschitz constants of the gradients, so that gradients that change quickly get sampled more often and gradients that change slowly get sampled less often \citep{schmidt2015non}, 
and ii) re-weighting the obtained gradients by inverses of their sampling probabilities to reduce the variance and unbias the estimates of gradients of the original objective function. 
The latter allows to make sure that all guarantees for the original and the re-weighted
updates are equally valid \citep{needell2014stochastic}.  

In contrast to these works, our approach does not re-weight the gradients and does not estimate their Lipschitz constants. 
This allows us to render the selection to be invariant to rank-preserving transformations of the loss function. 
To investigate whether our approach is viable outside its original domain of application, 
we empirically compared it to the Non-Uniform Stochastic Average Gradient Method (SAG-NUS*) by \cite{schmidt2015non}.  
SAG-NUS* is a state-of-the-art method which tracks estimates of Lipschitz constants to select training examples non-uniformly in order to speed up the solution of a strongly-convex optimization problem arising from the training of conditional random fields \citep{schmidt2015non}. Our batch selection approach showed competitive performance compared to SAG-NUS* (see Figure \ref{Sagres} in the supplementary material).

Non-uniform selection of batches of training data is virtually absent in deep learning. 
The only really closely relevant work we are aware of is a parallel ICLR 2016 submission by \cite{schaul2015prioritized}, 
where a very similar approach is used for acceleration of learning in Deep Q-Networks by prioritizing replay of relevant experiences. 
Another parallel ICLR submission by \cite{alain2015variance} employs importance sampling for variance reduction and to obtain an unbiased estimate of the gradient in the context of parallel evaluations of gradients; however, no non-uniform selection is used. 
From \cite{schaul2015prioritized}, we found that \cite{hinton2007recognize} used importance sampling for classification on MNIST (see also ``small-data'' experiments by \cite{simpson2015oddball}), and, interestingly, warned that the use of extreme values of selection probabilities such as 0.01 and 100 
is dangerous because ``a case that used to be easy might have developed a large gradient 
while it was being ignored, and multiplying this gradient by 100 could give the network
a shock''. Note that we do not re-weight gradients, which allows us to use much larger values for the difference in probabilities $s$, with a best-found value in the order of $10^8$. \cite{alain2015variance} also notice that their algorithm's stability is in danger due to the normalization by probabilities, the essential part of optimal importance sampling they employ (optimal in the sense of variance reduction in estimating the local gradient, but not necessarily in terms of better generalization, etc). To counter this effect, they add a small value to all probabilities, a parameter-dependent heuristic which may drift the final approach from its theoretically-sound origin. The authors of \citep{alain2015variance} note that their importance sampling approach is limited to fully connected networks (at least so far), whereas our approach directly works for any kind of network structure. 
We found that the use of importance sampling is not beneficial for MNIST when used with our batch selection approach (see Figure \ref{ISfig} in the supplementary material).
 
Curriculum learning is another relevant concept where the examples are not randomly presented but organized in a meaningful order which illustrates gradually more concepts, and gradually more complex ones \cite{bengio2009curriculum}. 
In the context of deep learning,  
curriculum learning can be viewed as a particular case of batch selection when $s<1$ and not $s>1$ as we explore in this paper. 

\section{Limitations}

\textbf{Time complexity}. The sorting procedure leads to a negligible overhead 
for MNIST with 50.000 training datapoints. Much larger datasets will increase 
the absolute overhead of the sorting but the relative overhead might continue 
being negligible compared to the cost of learning since networks are typically much bigger then. 
If this is not the case, the bisection algorithm for sorting might be used instead of the current naive use of QuickSort. The overhead of the whole procedure then would scale as $O(\log N )$ per datapoint. 
The dominant overhead comes from the recomputation procedure, but this can be controlled by 
$r_{freq}$ and $r_{ratio}$. 

\textbf{Hyperparameter settings}. 
While the values of $s$ in order of $10^2$ - $10^8$ seem reasonable for MNIST, 
it remains unclear what values should be used for other datasets. 
As for other hyperparameters, we can of course use automated hyperparameter optimization methods to 
set $s$, but we may also be able to derive a hyperparameter-less robust adaptive schedule.
The simplest approach to choose $s$ would be to set it to a small value, e.g., 
 $s=10$ and iteratively increase it, e.g., by a factor of 10 until 
the performance does not improve anymore. In this case, 
 $r_{ratio}$ might be set to 1 and $r_{freq}$ (initially set to 0) should be increased. 
It would be of interest to automate this process similarly to the online hyperparameter adaptation procedure given in \cite{loshchilov2014maximum}, where an internal success criterion is considered. 
While both AdaDelta and Adam are claimed to be robust to their hyperparameter settings, 
we found that their performance can be further improved by means of hyperparameter tuning (see Figure \ref{FigureMnistAll}). 
However, the question of optimal hyperparameter settings is of great interest and we plan to investigate it further with CMA-ES \citep{hansen2001completely} and standard Bayesian optimization methods~\citep{bartz2005sequential,snoek-nips12a,bergstra-nips11a,hutter-lion11a}.

\textbf{Generalization to other datasets}. 
The main limitation of the current work is the breadth of its empirical evaluation, which only considered training of deep networks for the MNIST dataset and training of conditional random fields for optimal character recognition. 
The proposed method is expected to work well provided that the ranks of datapoints w.r.t. loss values are relatively stable over time. Our preliminary experiments with the CIFAR-10 dataset so far only show benefits of batch selection over random selection and not over shuffling; we are actively working on analyzing the possible causes for this and at this point only report some preliminary experiments in the supplementary material (Section \ref{sec:hyperparam_opt}).

\section{Conclusion}

While the common approach to deal with the stochasticity involved 
in training of deep learning networks considers a post-processing of the gradient information incoming from batches of the dataset, we propose to control the source of this information, the batch itself.  
More specifically, we suggest to more frequently select training datapoints with the greatest contribution to the objective function. The probability of each datapoint to be selected is decreasing with the rank of its latest known loss value. Thus, the selection itself is invariant to rank-preserving transformation of the loss function. 

Our experiments demonstrated that online batch selection speeds up the convergence of the state-of-the-art DNN training methods AdaDelta and Adam by a factor of about 5. 
We also envision that a tighter coupling of learning approaches and batch selection would allow to provide additional speedups; for instance, one should properly link learning rates and batch size, arguably, two sides of the same coin. The increase of the batch size over time seems to be promising to deal with the noise involved in gradient estimation when approaching the optimum (see  theoretical investigations by \cite{friedlander2012hybrid,babanezhad2015stop} and section \ref{bsizesection} in the supplementary material).

This paper provides only an initial study of possible online batch selection strategies. 
More advanced approaches might involve not only objective function values but also some additional indicators, such as gradient amplitudes, datapoint features/classes, and similarity to other datapoints. The selection then may consider not only the training objective function but also some proxies to the generalization capabilities of the network.  

Overall, we showed that even simple batch selection mechanisms can lead to a substantial 
performance improvement. 
We hope that this finding will attract attention to this underexplored approach for
 speeding up the training of neural networks.

\bibliography{iclr2016_conference}

\begin{thebibliography}{39}
\providecommand{\natexlab}[1]{#1}
\providecommand{\url}[1]{\texttt{#1}}
\expandafter\ifx\csname urlstyle\endcsname\relax
  \providecommand{\doi}[1]{doi: #1}\else
  \providecommand{\doi}{doi: \begingroup \urlstyle{rm}\Url}\fi

\bibitem[Alain et~al.(2015)Alain, Lamb, Sankar, Courville, and
  Bengio]{alain2015variance}
Alain, Guillaume, Lamb, Alex, Sankar, Chinnadhurai, Courville, Aaron, and
  Bengio, Yoshua.
\newblock Variance reduction in sgd by distributed importance sampling.
\newblock \emph{arXiv preprint arXiv:1511.06481}, 2015.

\bibitem[Babanezhad et~al.(2015)Babanezhad, Ahmed, Virani, Schmidt,
  Kone{\v{c}}n{\`y}, and Sallinen]{babanezhad2015stop}
Babanezhad, Reza, Ahmed, Mohamed~Osama, Virani, Alim, Schmidt, Mark,
  Kone{\v{c}}n{\`y}, Jakub, and Sallinen, Scott.
\newblock Stop wasting my gradients: Practical svrg.
\newblock \emph{arXiv preprint arXiv:1511.01942}, 2015.

\bibitem[B{\"a}ck(1996)]{back1996evolutionary}
B{\"a}ck, Thomas.
\newblock \emph{Evolutionary algorithms in theory and practice: evolution
  strategies, evolutionary programming, genetic algorithms}.
\newblock Oxford university press, 1996.

\bibitem[Bartz-Beielstein et~al.(2005)Bartz-Beielstein, Lasarczyk, and
  Preu{\ss}]{bartz2005sequential}
Bartz-Beielstein, Thomas, Lasarczyk, Christian~WG, and Preu{\ss}, Mike.
\newblock Sequential parameter optimization.
\newblock In \emph{Evolutionary Computation, 2005. The 2005 IEEE Congress on},
  volume~1, pp.\  773--780. IEEE, 2005.

\bibitem[Bengio et~al.(2009)Bengio, Louradour, Collobert, and
  Weston]{bengio2009curriculum}
Bengio, Yoshua, Louradour, J{\'e}r{\^o}me, Collobert, Ronan, and Weston, Jason.
\newblock Curriculum learning.
\newblock In \emph{Proceedings of the 26th annual international conference on
  machine learning}, pp.\  41--48. ACM, 2009.

\bibitem[Bergstra et~al.(2011)Bergstra, Bardenet, Bengio, and
  K{\'e}gl]{bergstra-nips11a}
Bergstra, J., Bardenet, R., Bengio, Y., and K{\'e}gl, B.
\newblock Algorithms for hyper-parameter optimization.
\newblock In \emph{Proc. of NIPS'11}, pp.\  2546--2554, 2011.

\bibitem[Bordes et~al.(2009)Bordes, Bottou, and Gallinari]{bordes2009sgd}
Bordes, Antoine, Bottou, L{\'e}on, and Gallinari, Patrick.
\newblock Sgd-qn: Careful quasi-newton stochastic gradient descent.
\newblock \emph{The Journal of Machine Learning Research}, 10:\penalty0
  1737--1754, 2009.

\bibitem[Choromanska et~al.(2014)Choromanska, Henaff, Mathieu, Arous, and
  LeCun]{choromanska2014loss}
Choromanska, Anna, Henaff, Mikael, Mathieu, Michael, Arous, G{\'e}rard~Ben, and
  LeCun, Yann.
\newblock The loss surface of multilayer networks.
\newblock \emph{arXiv preprint arXiv:1412.0233}, 2014.

\bibitem[Courbariaux et~al.(2015)Courbariaux, Bengio, and
  David]{courbariaux2015binaryconnect}
Courbariaux, Matthieu, Bengio, Yoshua, and David, Jean-Pierre.
\newblock Binaryconnect: Training deep neural networks with binary weights
  during propagations.
\newblock \emph{arXiv preprint arXiv:1511.00363}, 2015.

\bibitem[Dauphin et~al.(2014)Dauphin, Pascanu, Gulcehre, Cho, Ganguli, and
  Bengio]{dauphin2014identifying}
Dauphin, Yann~N, Pascanu, Razvan, Gulcehre, Caglar, Cho, Kyunghyun, Ganguli,
  Surya, and Bengio, Yoshua.
\newblock Identifying and attacking the saddle point problem in
  high-dimensional non-convex optimization.
\newblock In \emph{Advances in Neural Information Processing Systems}, pp.\
  2933--2941, 2014.

\bibitem[Dauphin et~al.(2015)Dauphin, de~Vries, Chung, and
  Bengio]{dauphin2015rmsprop}
Dauphin, Yann~N, de~Vries, Harm, Chung, Junyoung, and Bengio, Yoshua.
\newblock Rmsprop and equilibrated adaptive learning rates for non-convex
  optimization.
\newblock \emph{arXiv preprint arXiv:1502.04390}, 2015.

\bibitem[Deng et~al.(2013)Deng, Hinton, and Kingsbury]{Deng13newtypes}
Deng, L., Hinton, G., and Kingsbury, B.
\newblock New types of deep neural network learning for speech recognition and
  related applications: An overview.
\newblock In \emph{Proc. of ICASSP'13}, 2013.

\bibitem[Donahue et~al.(2014)Donahue, Jia, Vinyals, Hoffman, Zhang, Tzeng, and
  Darrell]{Donahue_ICML2014}
Donahue, J., Jia, Y., Vinyals, O., Hoffman, J., Zhang, N., Tzeng, E., and
  Darrell, T.
\newblock Decaf: A deep convolutional activation feature for generic visual
  recognition.
\newblock In \emph{Proc. of ICML'14}, 2014.

\bibitem[Duchi et~al.(2011)Duchi, Hazan, and Singer]{duchi2011adaptive}
Duchi, John, Hazan, Elad, and Singer, Yoram.
\newblock Adaptive subgradient methods for online learning and stochastic
  optimization.
\newblock \emph{The Journal of Machine Learning Research}, 12:\penalty0
  2121--2159, 2011.

\bibitem[Fletcher(1970)]{fletcher1970new}
Fletcher, Roger.
\newblock A new approach to variable metric algorithms.
\newblock \emph{The computer journal}, 13\penalty0 (3):\penalty0 317--322,
  1970.

\bibitem[Friedlander \& Schmidt(2012)Friedlander and
  Schmidt]{friedlander2012hybrid}
Friedlander, Michael~P and Schmidt, Mark.
\newblock Hybrid deterministic-stochastic methods for data fitting.
\newblock \emph{SIAM Journal on Scientific Computing}, 34\penalty0
  (3):\penalty0 A1380--A1405, 2012.

\bibitem[Fukumizu \& Amari(2000)Fukumizu and Amari]{fukumizu2000local}
Fukumizu, Kenji and Amari, Shun-ichi.
\newblock Local minima and plateaus in hierarchical structures of multilayer
  perceptrons.
\newblock \emph{Neural Networks}, 13\penalty0 (3):\penalty0 317--327, 2000.

\bibitem[Hansen \& Ostermeier(2001)Hansen and Ostermeier]{hansen2001completely}
Hansen, Nikolaus and Ostermeier, Andreas.
\newblock Completely derandomized self-adaptation in evolution strategies.
\newblock \emph{Evolutionary computation}, 9\penalty0 (2):\penalty0 159--195,
  2001.

\bibitem[Hansen \& Ros(2010)Hansen and Ros]{hansen2010benchmarking}
Hansen, Nikolaus and Ros, Raymond.
\newblock Benchmarking a weighted negative covariance matrix update on the
  bbob-2010 noiseless testbed.
\newblock In \emph{Proceedings of the 12th annual conference companion on
  Genetic and evolutionary computation}, pp.\  1673--1680. ACM, 2010.

\bibitem[Hazan et~al.(2015)Hazan, Levy, and Shalev-Shwartz]{hazan2015beyond}
Hazan, Elad, Levy, Kfir~Y, and Shalev-Shwartz, Shai.
\newblock Beyond convexity: Stochastic quasi-convex optimization.
\newblock \emph{arXiv preprint arXiv:1507.02030}, 2015.

\bibitem[Hinton(2007)]{hinton2007recognize}
Hinton, Geoffrey~E.
\newblock To recognize shapes, first learn to generate images.
\newblock \emph{Progress in brain research}, 165:\penalty0 535--547, 2007.

\bibitem[Hutter et~al.(2011)Hutter, Hoos, and Leyton-Brown]{hutter-lion11a}
Hutter, F., Hoos, H., and Leyton-Brown, K.
\newblock Sequential model-based optimization for general algorithm
  configuration.
\newblock In \emph{Proc. of LION'11}, pp.\  507--523, 2011.

\bibitem[Ioffe \& Szegedy(2015)Ioffe and Szegedy]{ioffe2015batch}
Ioffe, Sergey and Szegedy, Christian.
\newblock Batch normalization: Accelerating deep network training by reducing
  internal covariate shift.
\newblock \emph{arXiv preprint arXiv:1502.03167}, 2015.

\bibitem[Kingma \& Ba(2014)Kingma and Ba]{kingma2014adam}
Kingma, Diederik and Ba, Jimmy.
\newblock Adam: A method for stochastic optimization.
\newblock \emph{arXiv preprint arXiv:1412.6980}, 2014.

\bibitem[Krizhevsky et~al.(2012)Krizhevsky, Sutskever, and
  Hinton]{Krizhevsky2012}
Krizhevsky, A., Sutskever, I., and Hinton, G.
\newblock Imagenet classification with deep convolutional neural networks.
\newblock In \emph{Proc. of NIPS'12}, pp.\  1097--1105, 2012.

\bibitem[LeCun et~al.(1998)LeCun, Bottou, Bengio, and
  Haffner]{lecun1998gradient}
LeCun, Yann, Bottou, L{\'e}on, Bengio, Yoshua, and Haffner, Patrick.
\newblock Gradient-based learning applied to document recognition.
\newblock \emph{Proceedings of the IEEE}, 86\penalty0 (11):\penalty0
  2278--2324, 1998.

\bibitem[Liu \& Nocedal(1989)Liu and Nocedal]{liu1989limited}
Liu, Dong~C and Nocedal, Jorge.
\newblock On the limited memory bfgs method for large scale optimization.
\newblock \emph{Mathematical programming}, 45\penalty0 (1-3):\penalty0
  503--528, 1989.

\bibitem[Loshchilov et~al.(2014)Loshchilov, Schoenauer, Sebag, and
  Hansen]{loshchilov2014maximum}
Loshchilov, Ilya, Schoenauer, Marc, Sebag, Michele, and Hansen, Nikolaus.
\newblock Maximum likelihood-based online adaptation of hyper-parameters in
  {CMA-ES}.
\newblock In \emph{Parallel Problem Solving from Nature--PPSN XIII}, pp.\
  70--79. Springer, 2014.

\bibitem[Needell et~al.(2014)Needell, Ward, and Srebro]{needell2014stochastic}
Needell, Deanna, Ward, Rachel, and Srebro, Nati.
\newblock Stochastic gradient descent, weighted sampling, and the randomized
  kaczmarz algorithm.
\newblock In \emph{Advances in Neural Information Processing Systems}, pp.\
  1017--1025, 2014.

\bibitem[Nesterov(2012)]{nesterov2012efficiency}
Nesterov, Yu.
\newblock Efficiency of coordinate descent methods on huge-scale optimization
  problems.
\newblock \emph{SIAM Journal on Optimization}, 22\penalty0 (2):\penalty0
  341--362, 2012.

\bibitem[Roller(2004)]{roller2004max}
Roller, Ben Taskar Carlos Guestrin~Daphne.
\newblock Max-margin markov networks.
\newblock \emph{Advances in neural information processing systems},
  16:\penalty0 25, 2004.

\bibitem[Schaul et~al.(2012)Schaul, Zhang, and LeCun]{schaul2012no}
Schaul, Tom, Zhang, Sixin, and LeCun, Yann.
\newblock No more pesky learning rates.
\newblock \emph{arXiv preprint arXiv:1206.1106}, 2012.

\bibitem[Schaul et~al.(2015)Schaul, Quan, Antonoglou, and
  Silver]{schaul2015prioritized}
Schaul, Tom, Quan, John, Antonoglou, Ioannis, and Silver, David.
\newblock Prioritized experience replay.
\newblock \emph{arXiv preprint arXiv:1511.05952}, 2015.

\bibitem[Schmidt et~al.(2015)Schmidt, Babanezhad, Ahmed, Defazio, Clifton, and
  Sarkar]{schmidt2015non}
Schmidt, Mark, Babanezhad, Reza, Ahmed, Mohamed~Osama, Defazio, Aaron, Clifton,
  Ann, and Sarkar, Anoop.
\newblock Non-uniform stochastic average gradient method for training
  conditional random fields.
\newblock \emph{arXiv preprint arXiv:1504.04406}, 2015.

\bibitem[Simpson(2015)]{simpson2015oddball}
Simpson, Andrew~JR.
\newblock " oddball sgd": Novelty driven stochastic gradient descent for
  training deep neural networks.
\newblock \emph{arXiv preprint arXiv:1509.05765}, 2015.

\bibitem[Snoek et~al.(2012)Snoek, Larochelle, and Adams]{snoek-nips12a}
Snoek, J., Larochelle, H., and Adams, R.~P.
\newblock Practical {B}ayesian optimization of machine learning algorithms.
\newblock In \emph{Proc. of NIPS'12}, pp.\  2960--2968, 2012.

\bibitem[Sutton \& Barto(1998)Sutton and Barto]{sutton1998reinforcement}
Sutton, Richard~S and Barto, Andrew~G.
\newblock \emph{Reinforcement learning: An introduction}, volume~1.
\newblock MIT press Cambridge, 1998.

\bibitem[Zeiler(2012)]{zeiler2012adadelta}
Zeiler, Matthew~D.
\newblock Adadelta: An adaptive learning rate method.
\newblock \emph{arXiv preprint arXiv:1212.5701}, 2012.

\bibitem[Zhao \& Zhang(2014)Zhao and Zhang]{zhao2014stochastic}
Zhao, Peilin and Zhang, Tong.
\newblock Stochastic optimization with importance sampling.
\newblock \emph{arXiv preprint arXiv:1401.2753}, 2014.

\end{thebibliography}
\bibliographystyle{iclr2016_conference}

\cleardoublepage

\section{Supplementary Material}

\subsection{When $N_{eff}=N$?} \label{beffb}

The selection pressure parameter $s$ affects the effective size of the training data we used. Let $N_{eff}$ denote the number of unique datapoints selected in the last $N$ steps; then, $N_{eff}$ is likely to decrease when $s$ increases. However, this is not necessarily the case: albeit unlikely, one possible scenario in which $N_{eff}$ would still equal $N$ is the following: if $p_1 = \infty$ and $p_{i>1}=0$, the datapoint with the greatest latest known loss $f_t(\vc{x}_t)$ is selected and the optimization algorithm $\A$ uses $\nabla f_t(\vc{x}_t)$ to minimize $f_t(\vc{x}_t)$, the new value $f_t(\vc{x}_t)$ appears to be the lowest loss value seen so far, the datapoint is ranked last and its selection is then scheduled to be in $N$ steps, the whole situation happens again and again leading to the sweep over the whole dataset and thus $N_{eff}=N$. 

\subsection{Online selection of the batch size}\label{bsizesection}

We decided to move the following material to simplify the main paper.

Not only the content of the selected batch, but also its size represents a design parameter of interest. When $b=N$, we have a denoised access to $f$. When $b=1$, we obtain a noisy but $N$ times cheaper access to $f$. The negative impact of this noise supposedly depends on the distance to the optimum in the objective and/or decision space. As we progress to the optimum over time, increasing the batch size is an approach to deal with the noise. Moreover, the use of a relatively small batch size in the beginning of search has a potential to bring a significant speedup, however, it has a potential risk of performing unnecessary random walk. 
We therefore propose to increase batch size over time. More specifically, we propose to consider both exponential and linear increases of $b$ over epochs $e$:

\begin{eqnarray}
	\label{eq:ev2}
	b_e=b_{e_0} \exp(\log(b_{e_{end}}/b_{e_0})/({e_{end}-{e_0}}))^{e_{end}-{e_0}},\\
	b_e=b_{e_0} + (b_{e_{end}} - b_{e_0}) (e-{e_0})/e_{end},
\end{eqnarray}

where ${e_0}$ and $e_{end}$ correspond to the first and last epochs of the increase (not necessarily the last epoch of optimization).

The closer approach of the optimum requires a greater batch size to be used to overcome the imprecision of gradient estimation. Therefore, starting with relatively small batches on easy training/optimization problems does not seem to be a bad idea. 
This intuition is supported by the recent results of \citeauthor{hazan2015beyond} showing that the Normalized Stochastic Gradient Descent (NSGD) (when the gradient is normalized by its amplitude) provably requires a minimum batch size to achieve an $\epsilon$ precision, without which the convergence of the NSGD cannot be guaranteed. 
The detailed theoretical analysis of advantages of batch size scheduling can be found in \citep{friedlander2012hybrid,babanezhad2015stop}.

We do not report the results for this section because the additional speedup provided by this procedure is 
pronounced in terms of epochs but not in terms of time. 
This is the case because small batches are more expensive (per datapoint) to compute on a GPU. 
However, we do include batch size selection procedure when dealing with hyper-parameter optimization 
(see section \ref{sec:hyperparam_opt}). 


\subsection{Importance sampling}

\begin{figure*}[t]
\begin{center}
\includegraphics[width=0.495\textwidth]{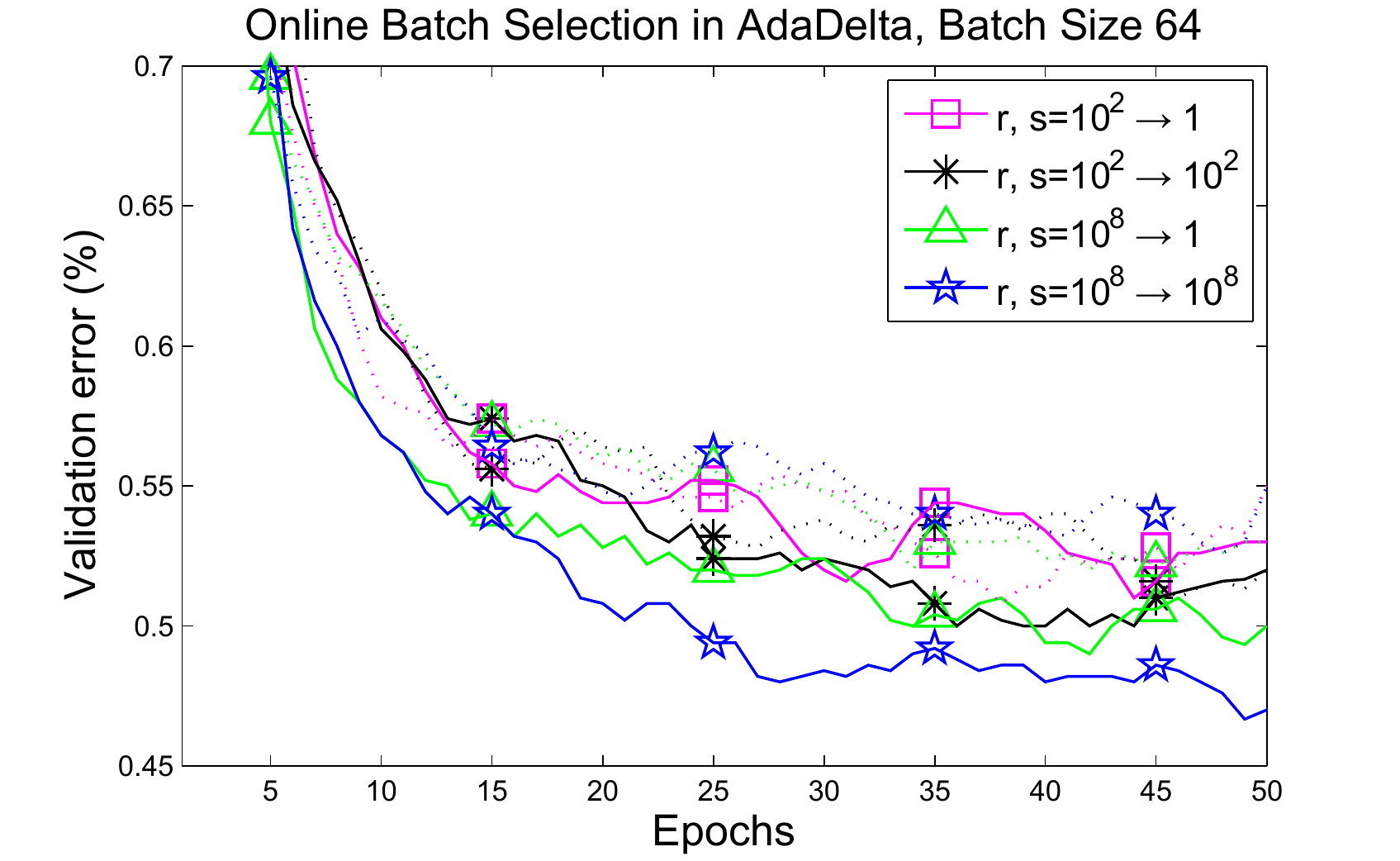}
\includegraphics[width=0.495\textwidth]{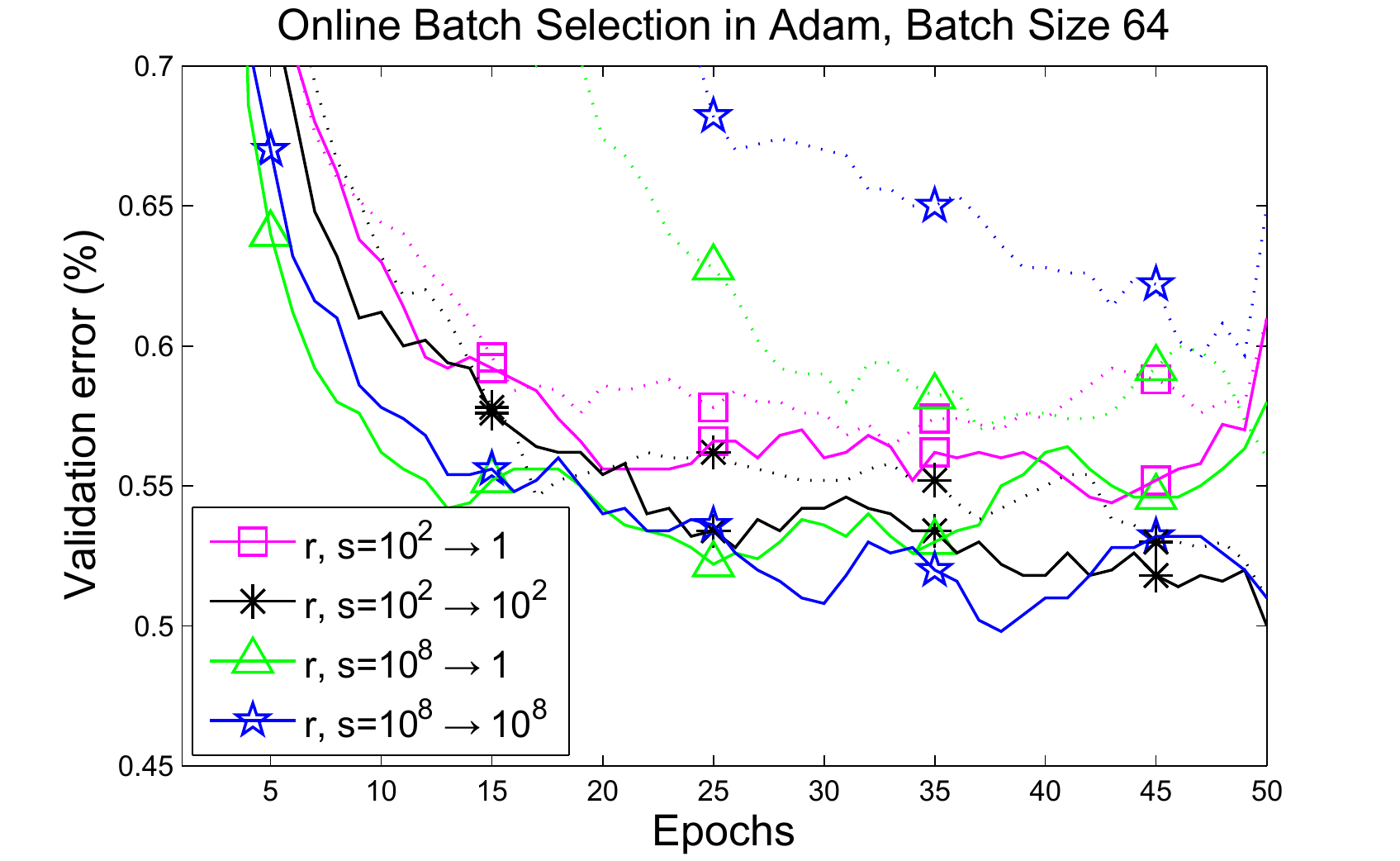}
\end{center}
\caption{
Convergence curves of AdaDelta (\textbf{Left}) and Adam (\textbf{Right}) on MNIST dataset. 
Bold lines correspond to our original approach, dotted lines correspond to its version with Importance Sampling.  
The value of $s$ denotes the ratio of probabilities to select the  training datapoint with the greatest latest known loss rather than the smallest latest known loss. 
Legends with 
$s=s_{e_0} \rightarrow s_{e_{end}}$ correspond to an exponential change of $s$ from $s_{e_0}$ to $s_{e_{end}}$ as a function of epoch index $e$, see Eq. (\ref{eq:ev1}). 
Legends with the prefix ``r'' correspond to the case when the loss values for $r_{ratio} \cdot N=1.0N$ datapoints with the greatest latest known loss 
 are recomputed $r_{freq}=0.5$ times per epoch. 
All curves are computed on the whole training set and smoothed by a moving average of the median of 11 runs.
}
\label{ISfig}
\end{figure*}

We use Importance Sampling (IS) on top of our batch selection approach. 
More specifically, we normalized gradients by normalized probabilities $p_{i}/\sum_{i=1}^{b} p_{i}$ of the corresponding 
datapoint to be selected~\citep{zhao2014stochastic}. 
Figure \ref{ISfig} shows the results of our original approach (solid lines) and its modified version (dotted lines) 
where IS is used. The latter shows a worse validation accuracy in all cases. 

\begin{figure*}[t]
\begin{center}
\includegraphics[width=0.495\textwidth]{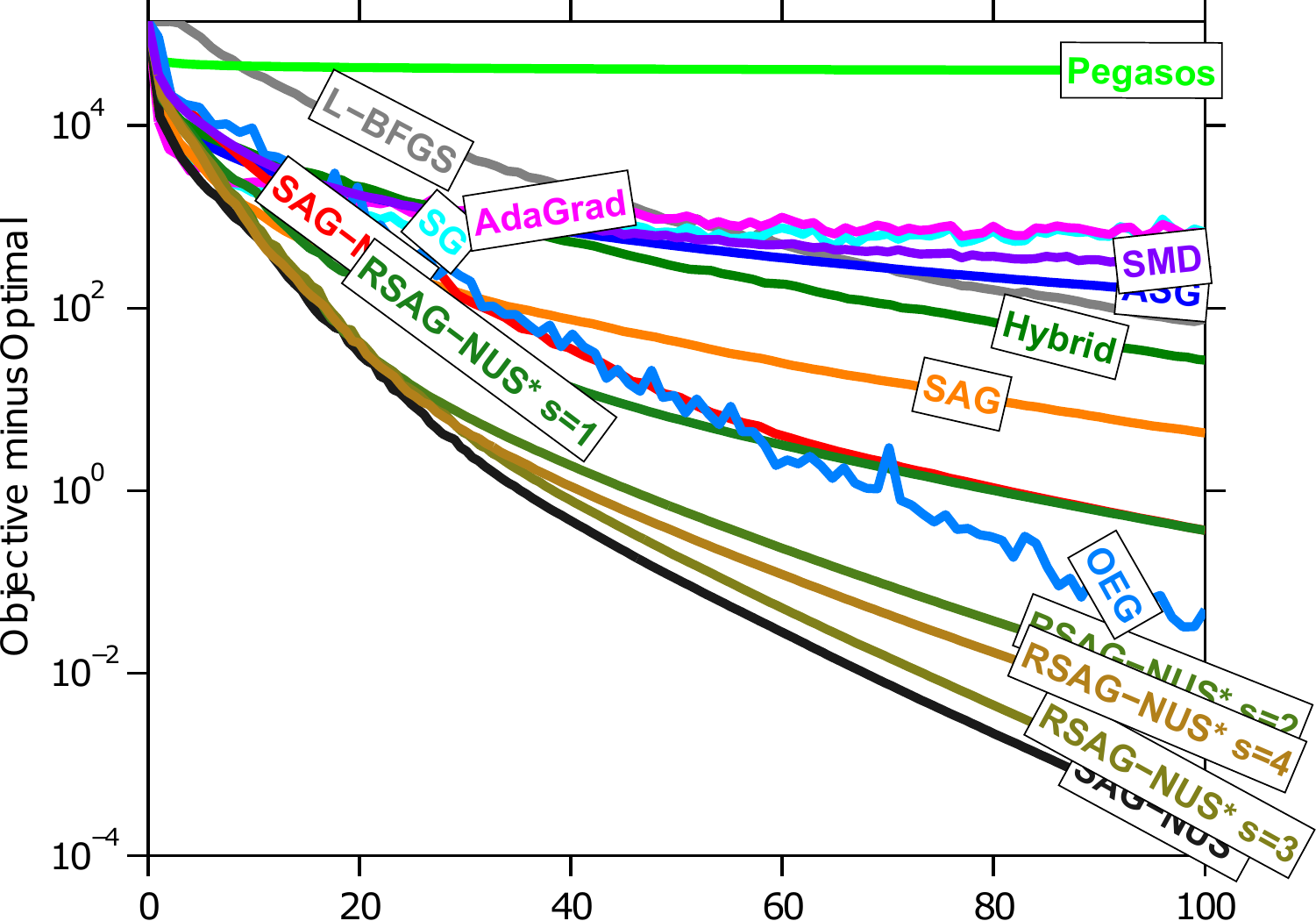}
\includegraphics[width=0.495\textwidth]{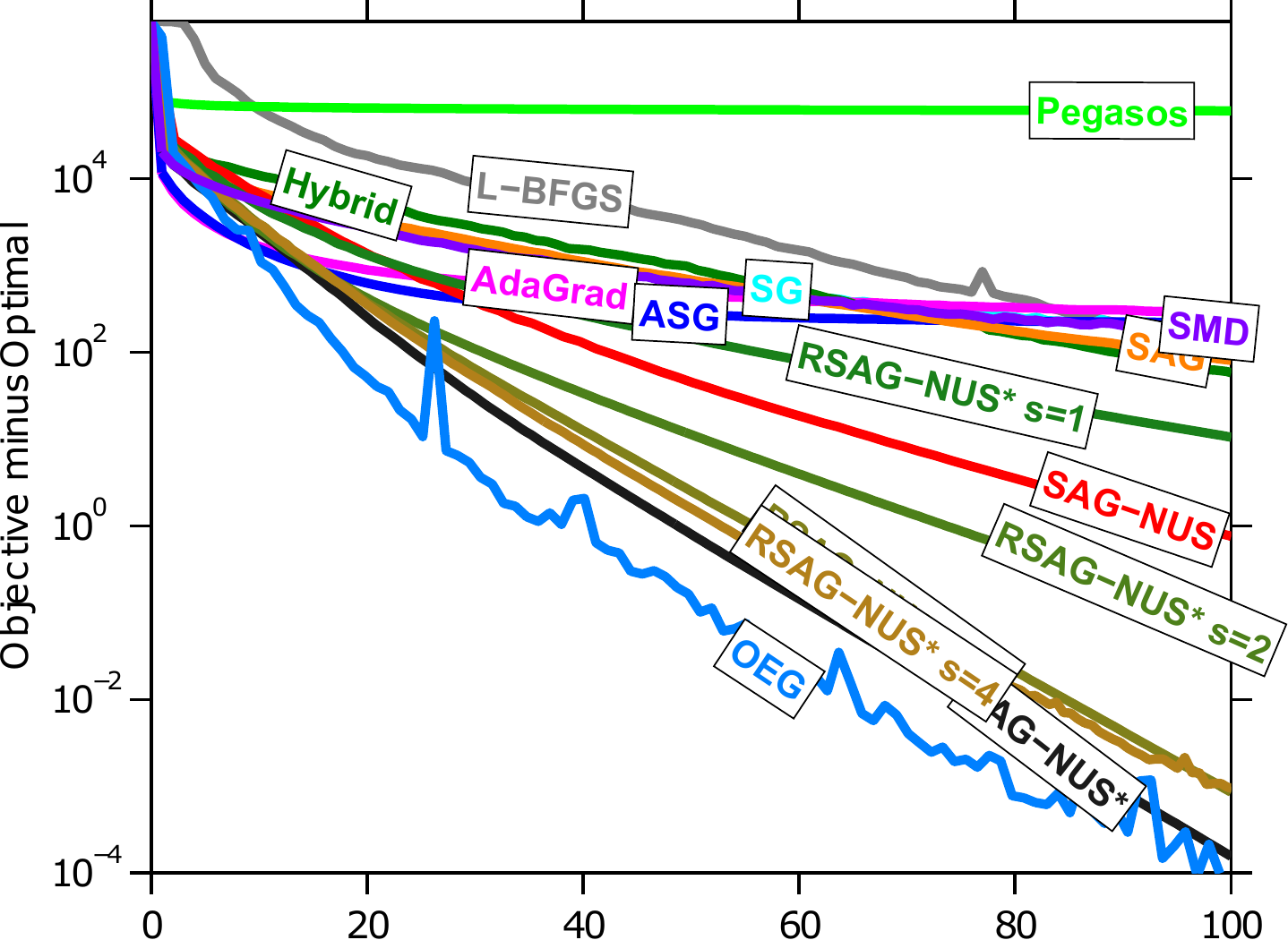}\\
\includegraphics[width=0.495\textwidth]{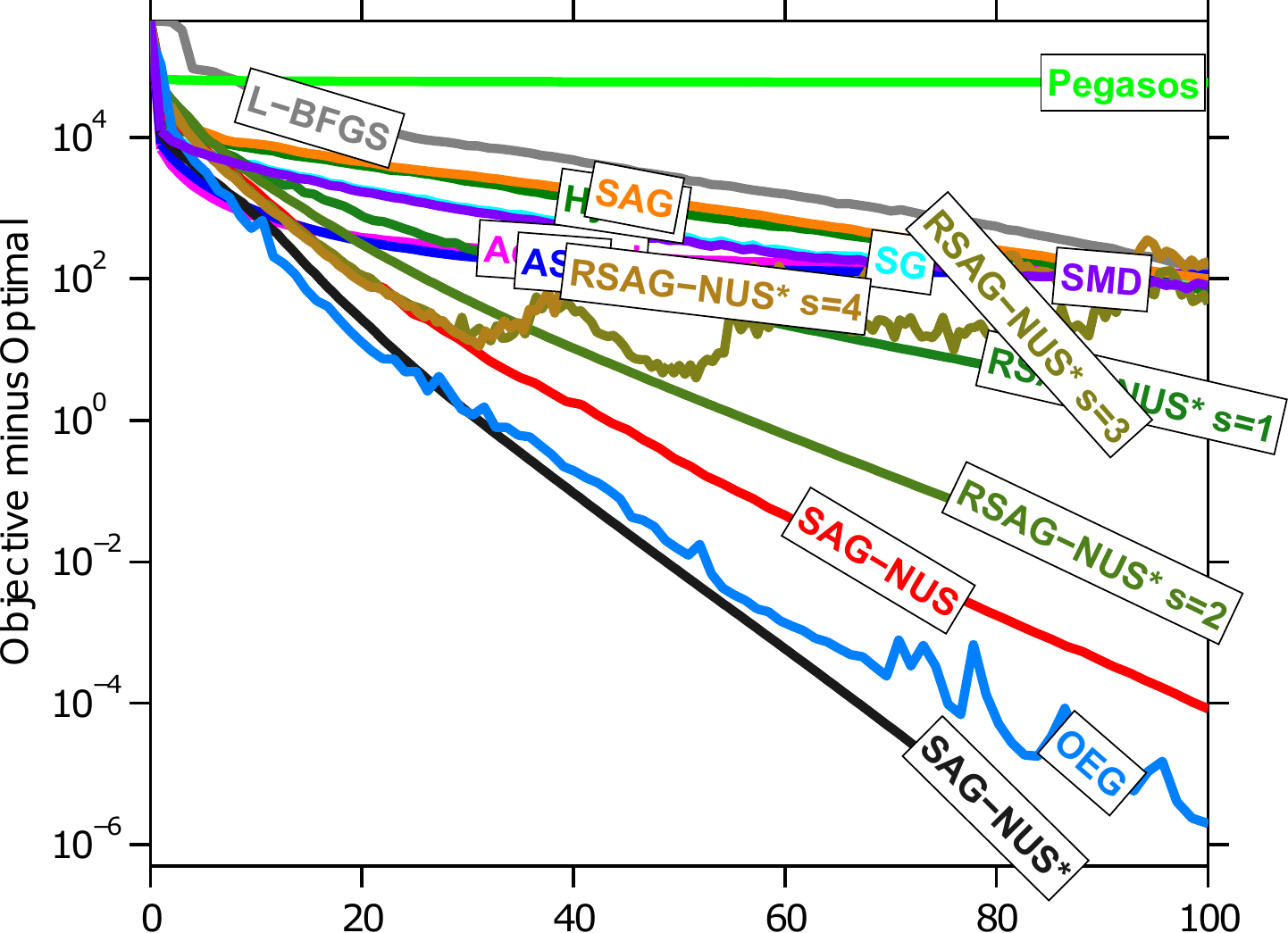}
\includegraphics[width=0.495\textwidth]{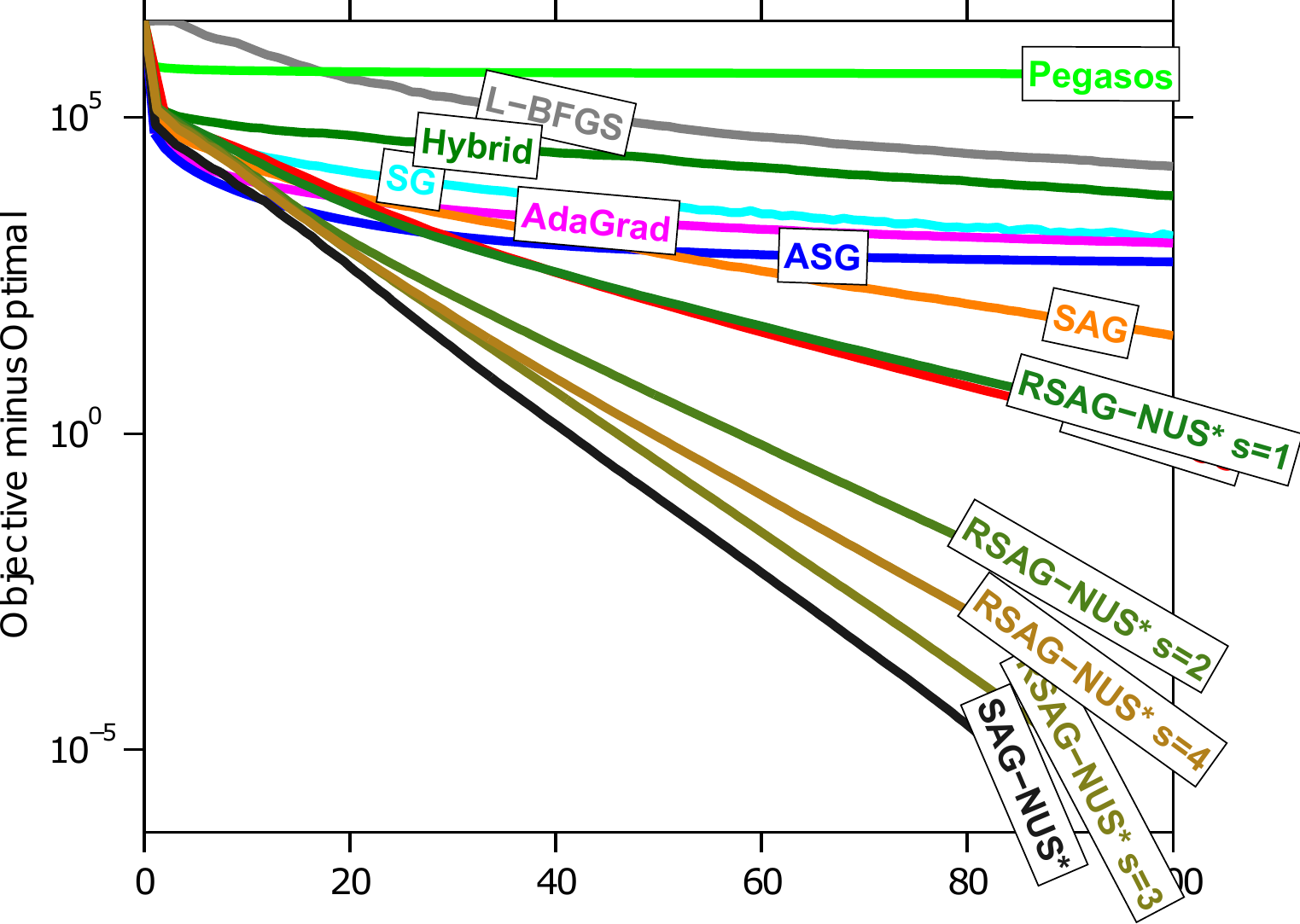}
\end{center}
\caption{
Objective minus optimal objective value against effective number of passes for different deterministic,
stochastic, and semi-stochastic optimization strategies. Top-left: OCR, Top-right: CoNLL-2000,
bottom-left: CoNLL-2002, bottom-right: POS-WSJ. 
}
\label{Sagres}
\end{figure*}

\subsection{Rank-based selection for Non-Uniform Stochastic Average Gradient Method}

The Non-Uniform Stochastic Average Gradient Method (SAG-NUS*) is a   
state-of-the-art method which tracks Lipschitz constants estimates to select 
training examples for faster solving a strongly-convex optimization problem arising from the training of Conditional Random Fields (CRFs) \citep{schmidt2015non}. 
We replaced the Lipschitz constants-based proportional selection by our rank-based selection which does not depend of Lipschitz constants themself but only on their ranks; 
the rest of SAG-NUS* remained unchanged. 
We benchmark a wide variety of approaches using the source code kindly provided by \cite{schmidt2015non} on four CRF training tasks: the optical character recognition
(OCR) dataset of \cite{roller2004max}, the CoNLL-2000 shallow parse chunking dataset 
\footnote{\url{http://www.cnts.ua.ac.be/conll2000/chunking}}, 
the CoNLL-2002
Dutch named-entity recognition dataset \footnote{\url{http://www.cnts.ua.ac.be/conll2002/ner}}, 
and a part-of-speech (POS) tagging task using the Penn Treebank Wall Street Journal data (POS-WSJ). 
The proper description of all algorithms and experimental settings is given in \citep{schmidt2015non}.  
We denote our modification of SAG-NUS* with rank-based selection as rSAG-NUS* and parameterize it 
by $s$ denoting the ratio between the greatest and smallest selection probabilities in the 
same way as in the main paper. We did not try to refresh the ranking as in the main paper. 

The results of rSAG-NUS* with $s=1$ showed in Figure \ref{Sagres} corresponds to uniform selection and defines a baseline. The results for rSAG-NUS* with $s=3$ are competitive to SAG-NUS* on three out of four datasets. 
The reason of a divergence seen for the CoNLL-2002 and rSAG-NUS* with $s=3$, $s=4$ is not yet clear  but 
the recomputation procedure (which we omitted in these experiments) might help to avoid it. 
This preliminary experiment suggests that our rank-based selection is a viable alternative (in three cases out of four) to the Lipschitz proportional selection. 
The source code to reproduce the experiments and Figure \ref{Sagres} is available at \url{https://sites.google.com/site/batchsel/}.

 \begin{table}[ht]
\caption{Hyperparameters descriptions, pseudocode transformations and ranges} 
\centering
\begin{tabular}{l*{6}{l}r}
name   	& description & transformation & range  \\
\hline
$x_1$ 	& selection pressure at ${e_0}$ & $10^{-2 + 10^{2 x_1}}$   & $[10^{-2}, 10^{98}]$  \\ 
$x_2$ 	& selection pressure at ${e_{end}}$ & $10^{-2 + 10^{2 x_2}}$   & $[10^{-2}, 10^{98}]$  \\
$x_3$ 	& batch size at ${e_{0}}$ & $2^{4 + 4 x_3}$   & $[2^4, 2^8]$  \\
$x_4$ 	& batch size at ${e_{end}}$ & $2^{4 + 4 x_4}$   & $[2^4, 2^8]$  \\
$x_5$ 	& frequency of loss recomputation $r_{freq}$ & $2 x_5$   & $[0, 2]$  \\
$x_6$ 	& alpha for batch normalization & $0.01 + 0.2 x_6$   & $[0.01, 0.21]$  \\
$x_7$ 	& epsilon for batch normalization & $10^{-8 + 5 x_7}$  & $[10^{-8}, 10^{-3}]$  \\
$x_8$ 	& dropout rate after the first Max-Pooling layer & $0.8 x_8$  & $[0, 0.8]$  \\
$x_9$ 	& dropout rate after the second Max-Pooling layer & $0.8 x_9$  & $[0, 0.8]$  \\
$x_{10}$ 	& dropout rate before the output layer & $0.8 x_{10}$  & $[0, 0.8]$  \\
$x_{11}$ 	& number of filters in the first convolution layer & $2^{3 + 5 x_{11}}$  & $[2^3, 2^8]$  \\
$x_{12}$ 	& number of filters in the second convolution layer & $2^{3 + 5 x_{12}}$  & $[2^3, 2^8]$  \\
$x_{13}$ 	& number of units in the fully-connected layer & $2^{4 + 5 x_{13}}$  & $[2^4, 2^9]$  \\
$x_{14}$ 	& Adadelta: learning rate at ${e_{0}}$ & $10^{0.5 - 2 x_{14}}$  & $[10^{-1.5}, 10^{0.5}]$  \\
$x_{15}$ 	& Adadelta: learning rate at ${e_{end}}$ & $10^{0.5 - 2 x_{15}}$  & $[10^{-1.5}, 10^{0.5}]$  \\
$x_{16}$ 	& Adadelta: $\rho$ & $0.8 + 0.199 x_{16}$  & $[0.8, 0.999]$  \\
$x_{17}$ 	& Adadelta: $\epsilon$ & $10^{-3 - 6 x_{17}}$  & $[10^{-9}, 10^{-3}]$  \\
$x_{14}$ 	& Adam: learning rate at ${e_{0}}$ & $10^{-1 - 3 x_{14}}$  & $[10^{-4}, 10^{-1}]$  \\
$x_{15}$ 	& Adam: learning rate at ${e_{end}}$ & $10^{-3 - 3 x_{15}}$  & $[10^{-6}, 10^{-3}]$  \\
$x_{16}$ 	& Adam: $\beta_1$ & $0.8 + 0.199 x_{16}$  & $[0.8, 0.999]$  \\
$x_{17}$ 	& Adam: $\epsilon$ & $10^{-3 - 6 x_{17}}$  & $[10^{-9}, 10^{-3}]$  \\
$x_{18}$ 	& Adam: $\beta_2$ & $1 - 10^{-2 - 2 x_{18}}$  & $[0.99, 0.9999]$  \\
$x_{19}$ 	& adaptation end epoch index ${e_{end}}$ & $20 + 200 x_{19}$  & $[20, 220]$  \\
\end{tabular}
\label{table:table} 
\end{table}

\begin{figure*}[p]
\begin{center}
\includegraphics[width=0.495\textwidth]{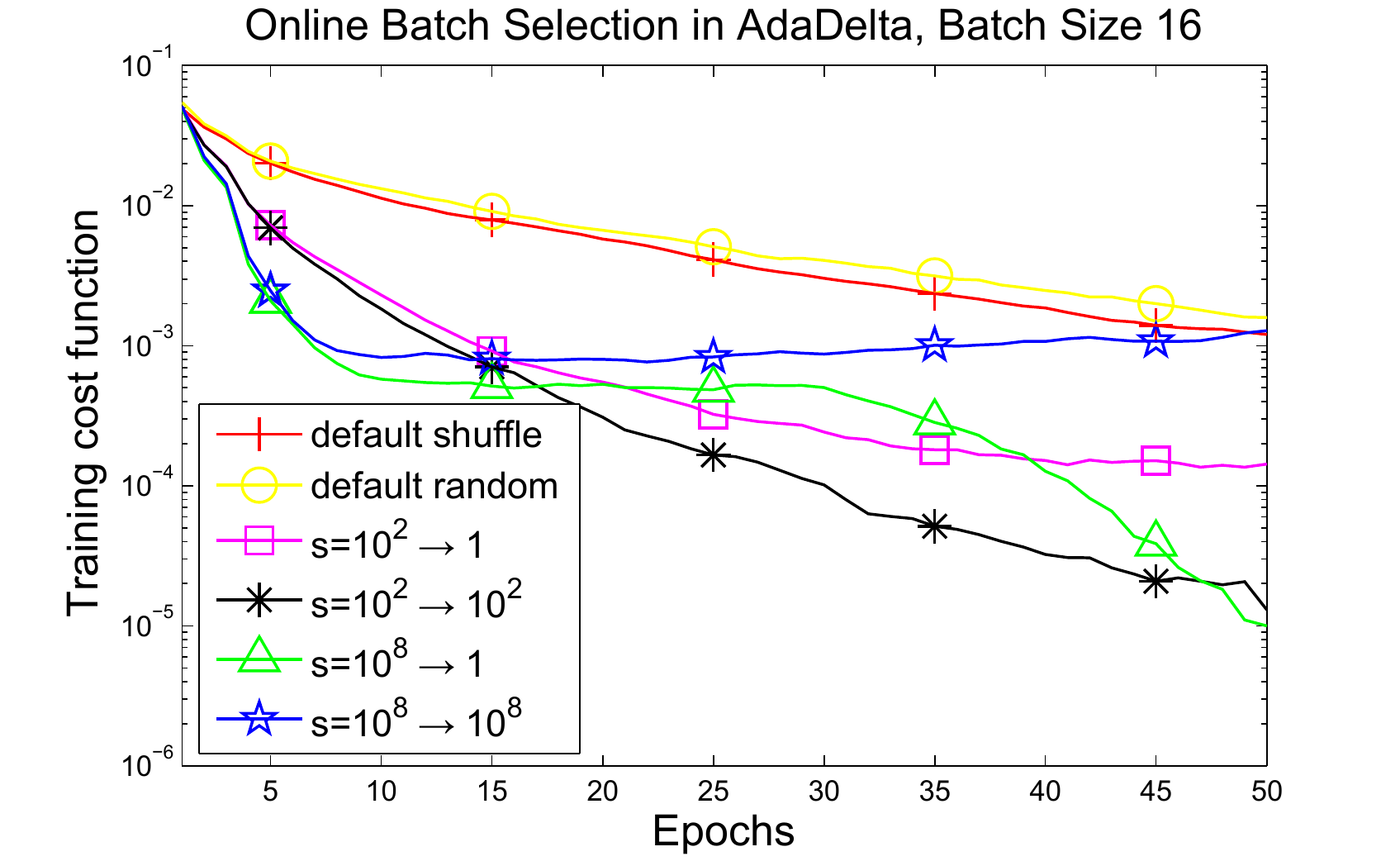}
\includegraphics[width=0.495\textwidth]{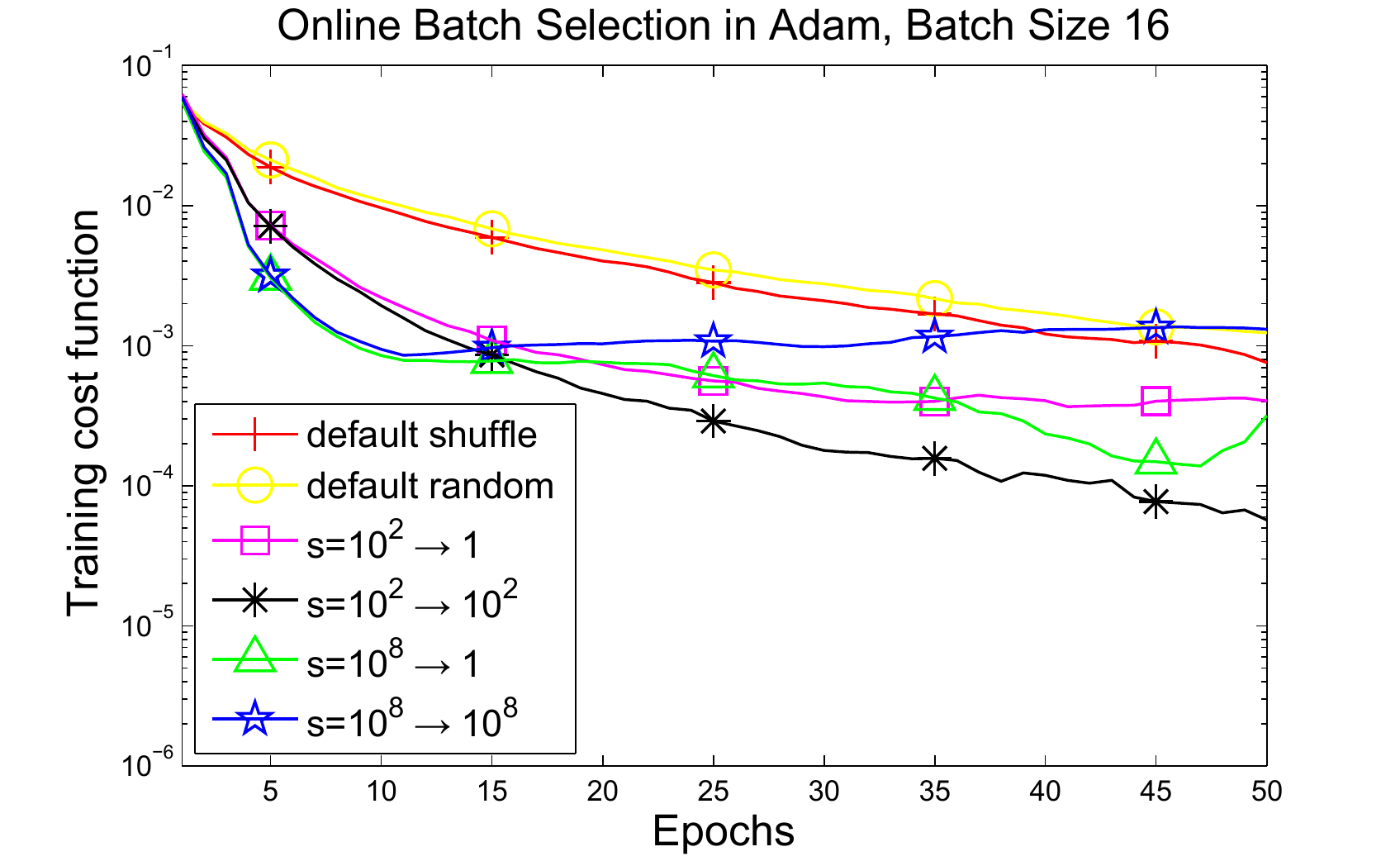}\\
\includegraphics[width=0.495\textwidth]{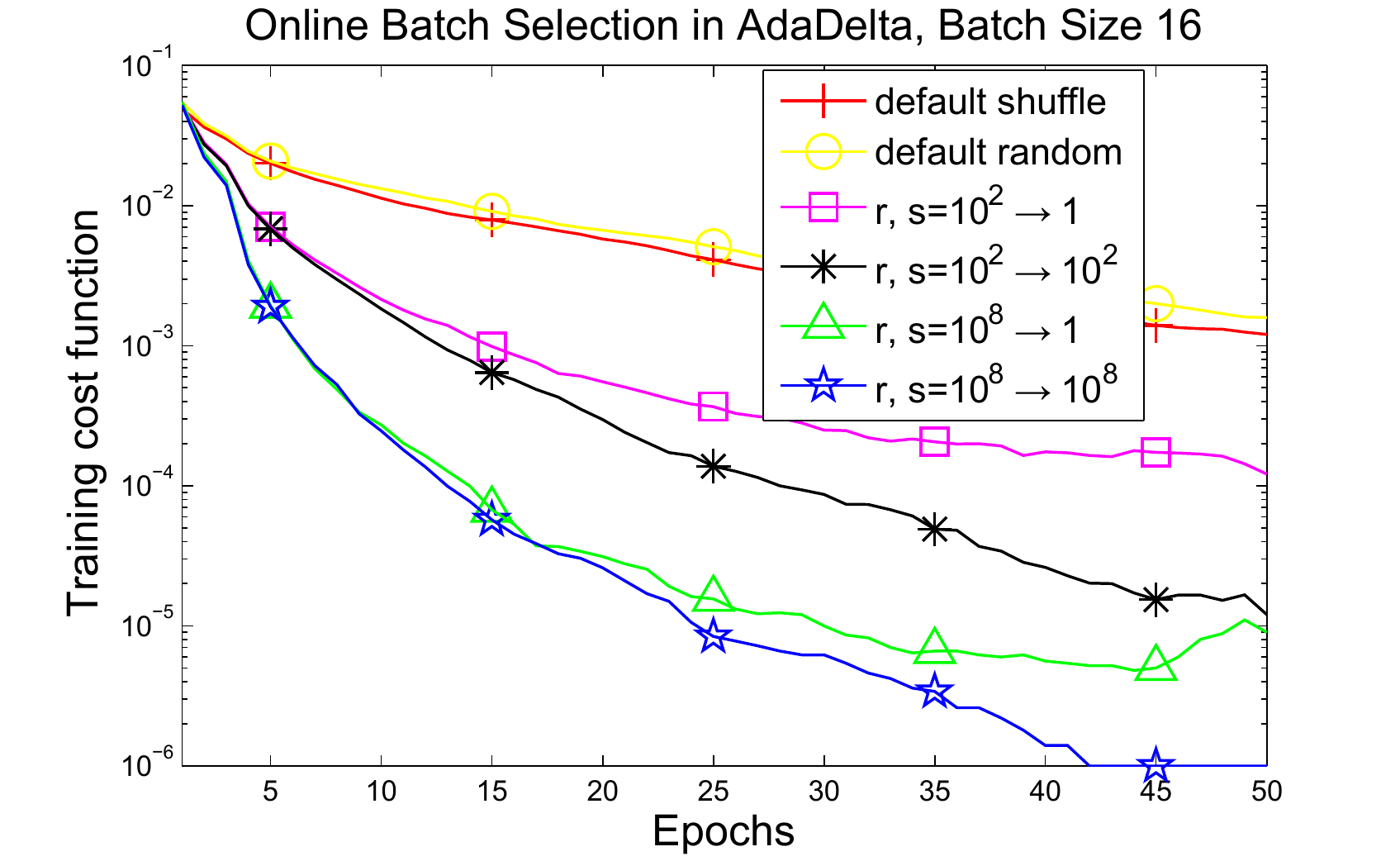}
\includegraphics[width=0.495\textwidth]{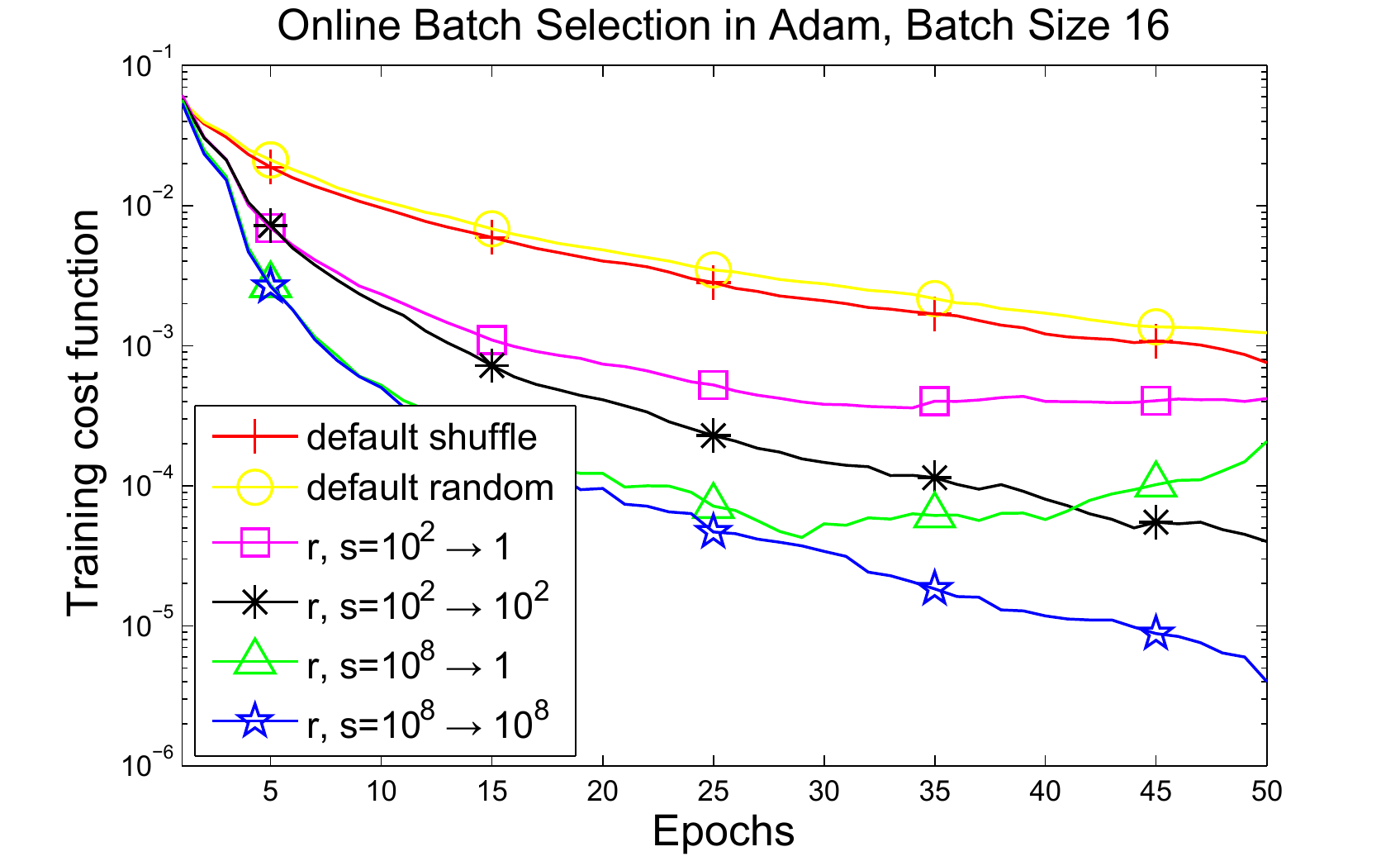}\\
\includegraphics[width=0.495\textwidth]{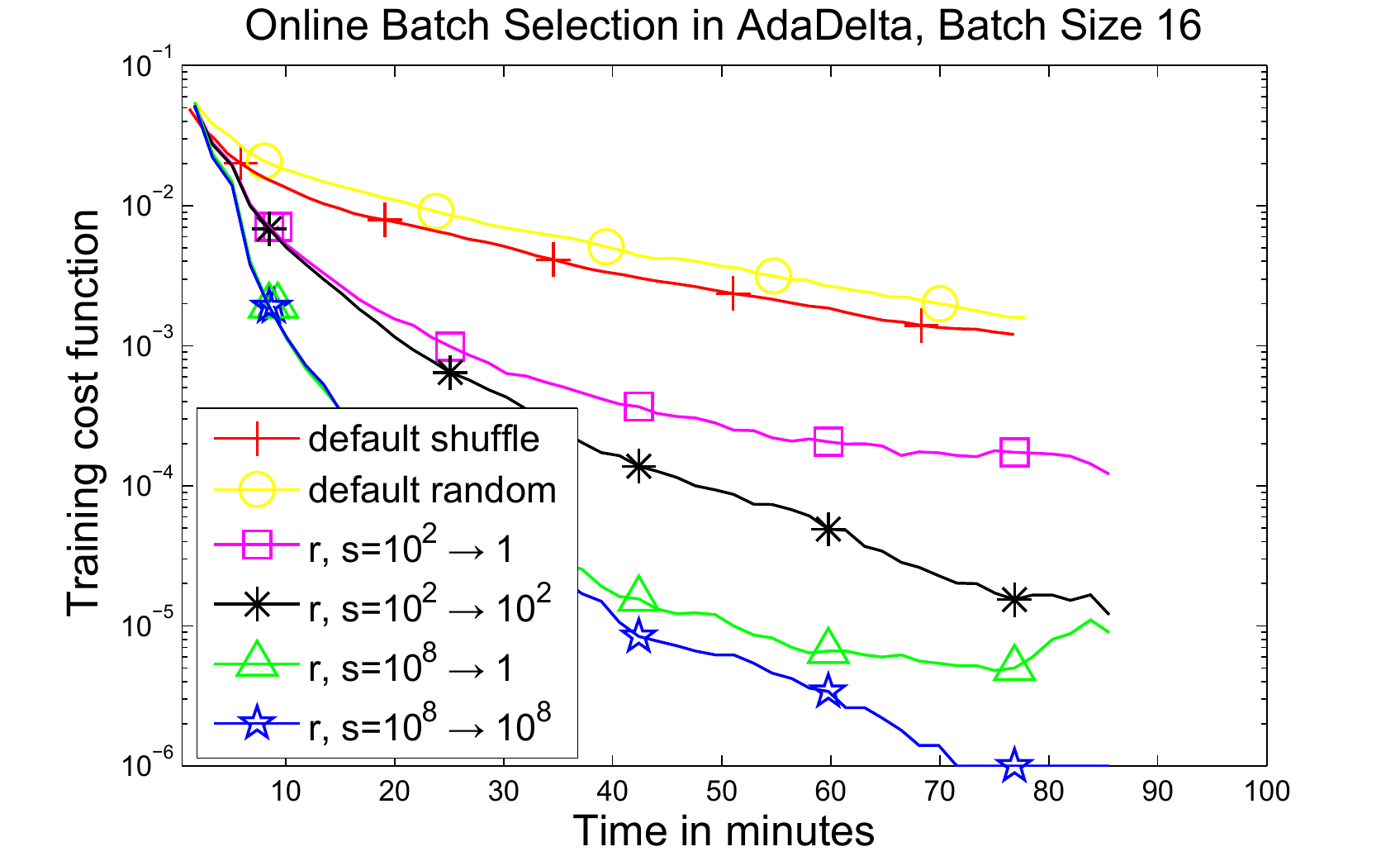}
\includegraphics[width=0.495\textwidth]{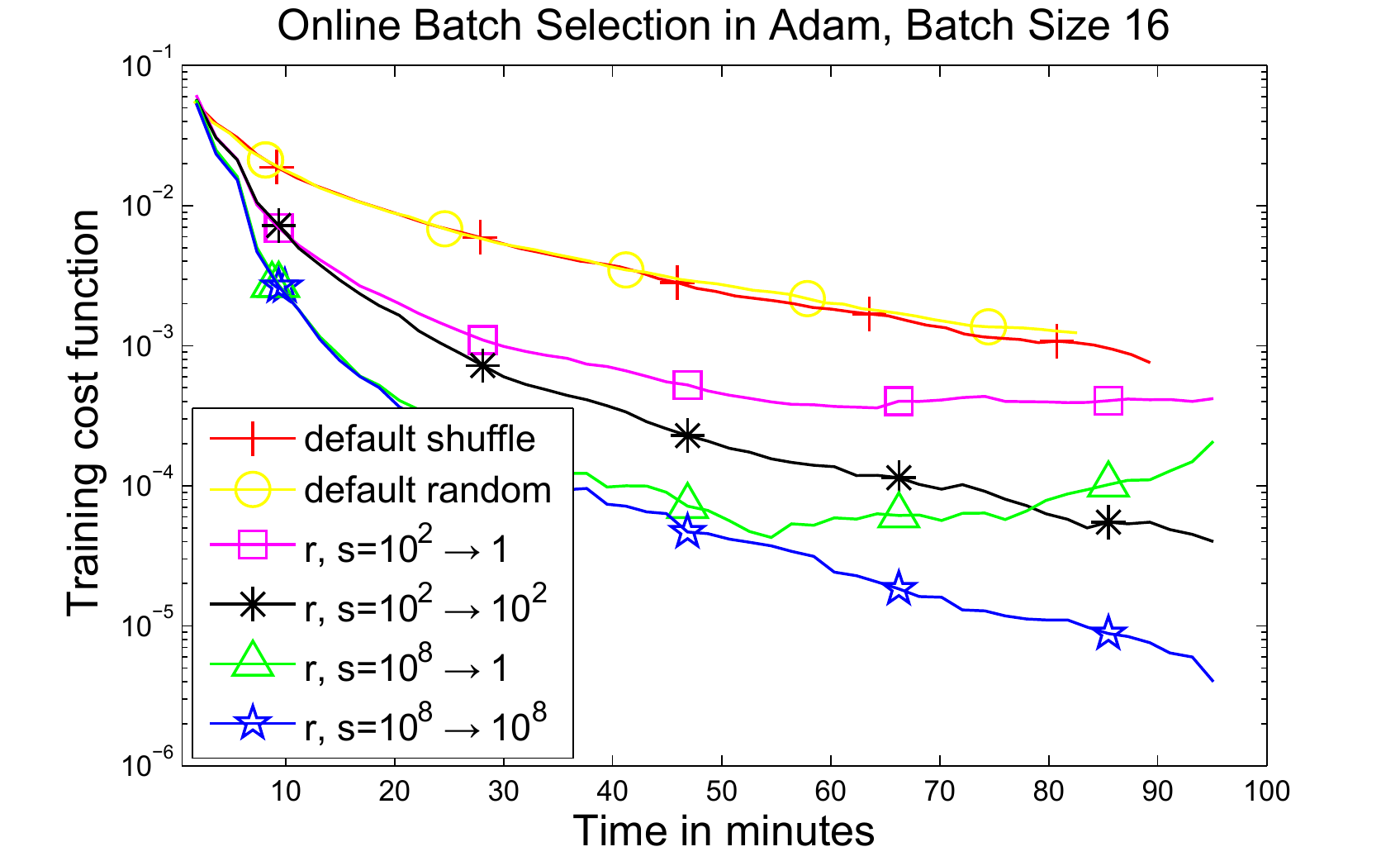}\\
\includegraphics[width=0.495\textwidth]{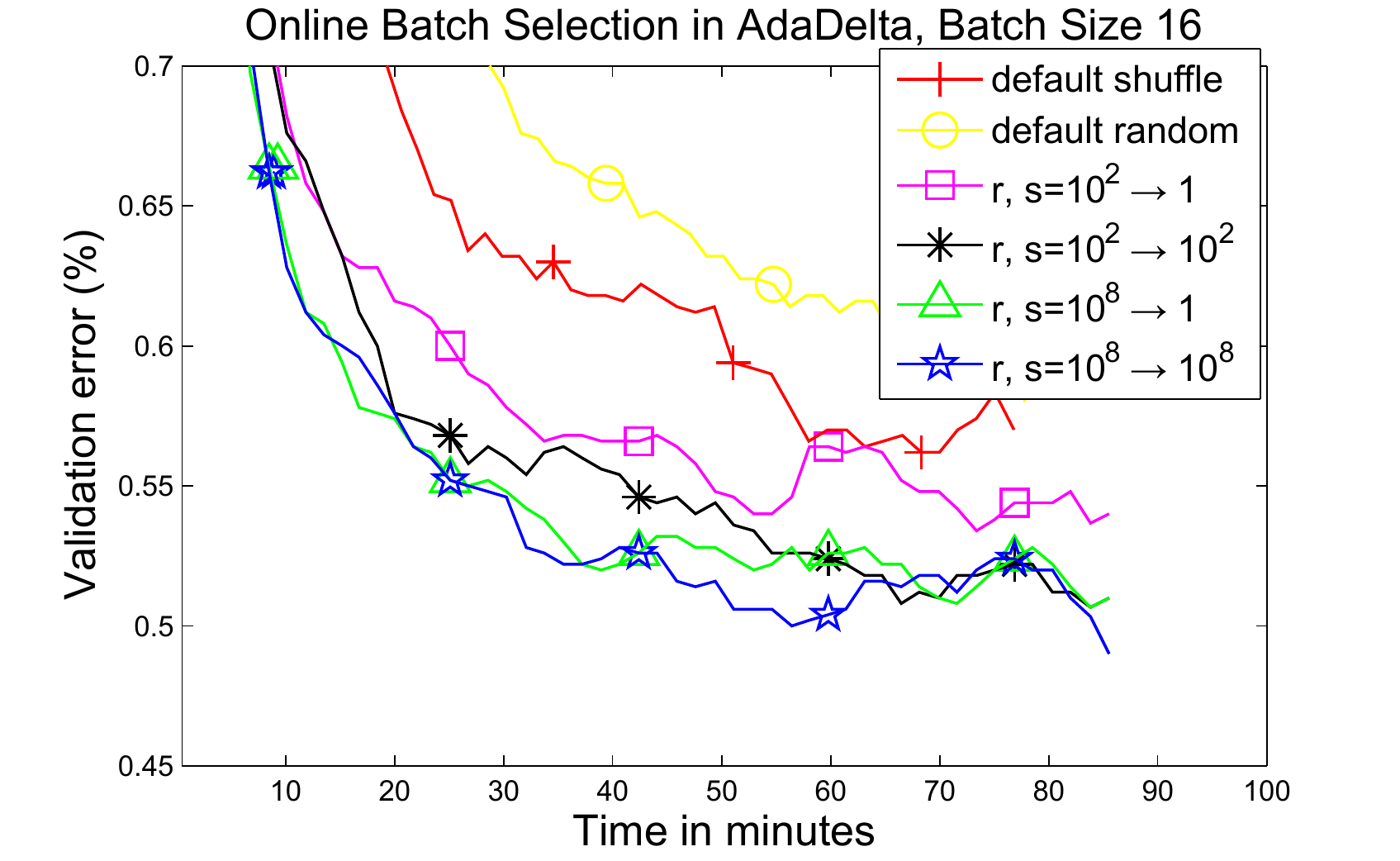}
\includegraphics[width=0.495\textwidth]{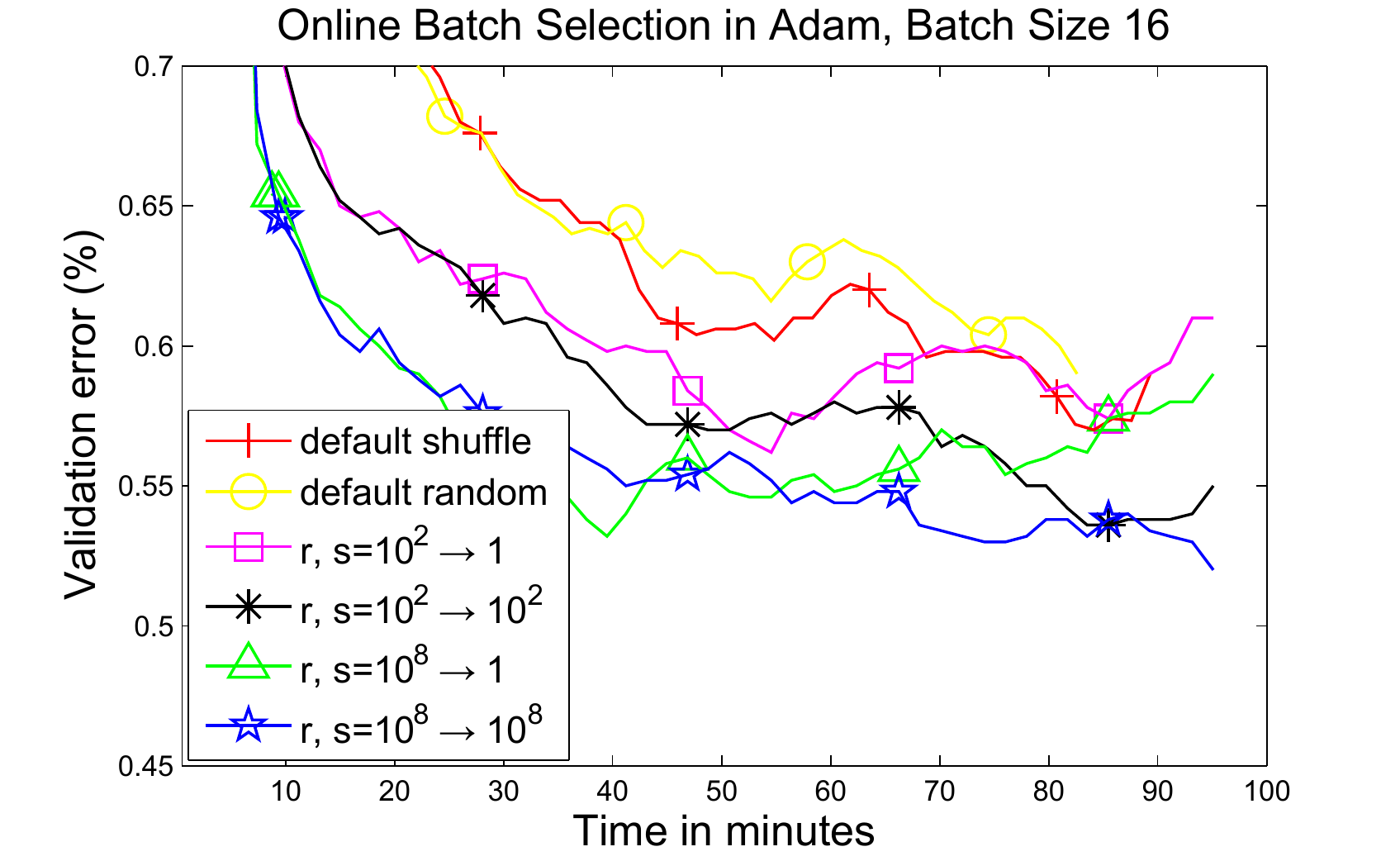}\\
\end{center}
\caption{
Convergence curves of AdaDelta (\textbf{Left Column}) and Adam (\textbf{Right Column}) on MNIST dataset. 
The original algorithms are denoted by ``default shuffle'' (respectively, ``default with random'') when datapoints are shuffled and selected sequentially (respectively, uniformly at random). 
The value of $s$ denotes the ratio of probabilities to select the  training datapoint with the greatest latest known loss rather than the smallest latest known loss. 
Legends with 
$s=s_{e_0} \rightarrow s_{e_{end}}$ correspond to an exponential change of $s$ from $s_{e_0}$ to $s_{e_{end}}$ as a function of epoch index $e$, see Eq. (\ref{eq:ev1}). 
Legends with the prefix ``r'' correspond to the case when the loss values for $r_{ratio} \cdot N=1.0N$ datapoints with the greatest latest known loss 
 are recomputed $r_{freq}=0.5$ times per epoch. 
All curves are computed on the whole training set and smoothed by a moving average of the median of 11 runs.
}
\label{Figure2}
\end{figure*}

\begin{figure*}[p]
\begin{center}
\includegraphics[width=0.495\textwidth]{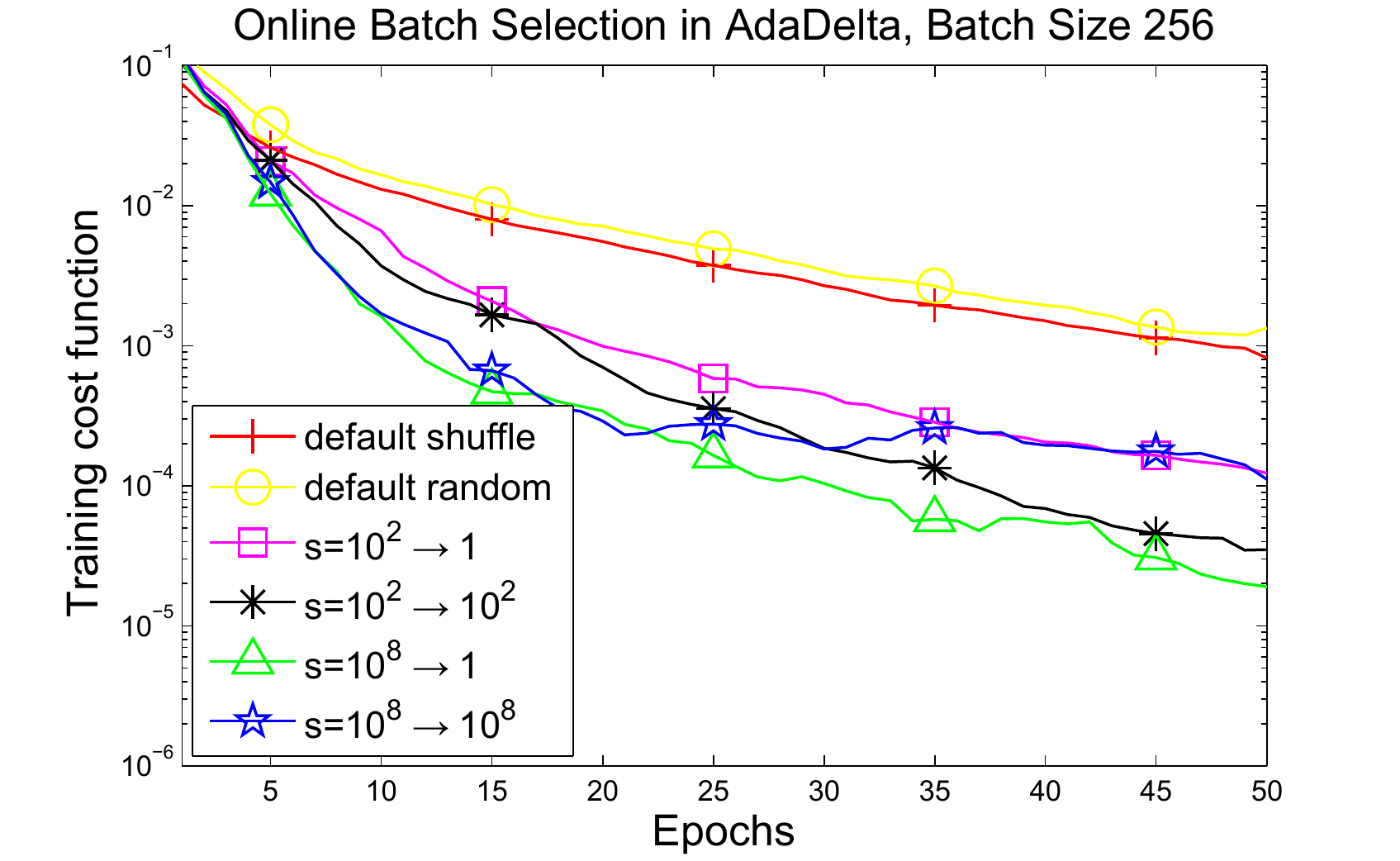}
\includegraphics[width=0.495\textwidth]{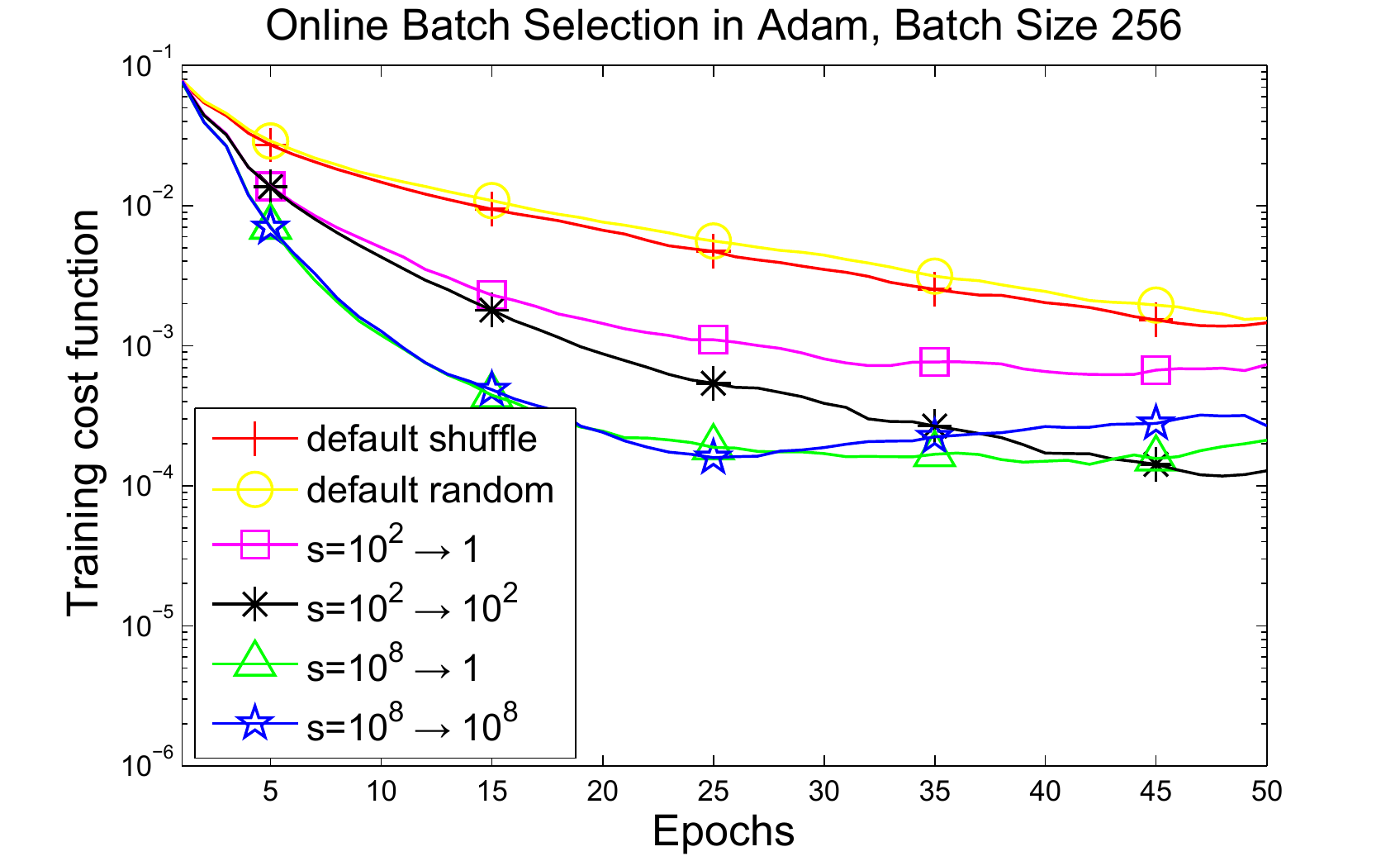}\\
\includegraphics[width=0.495\textwidth]{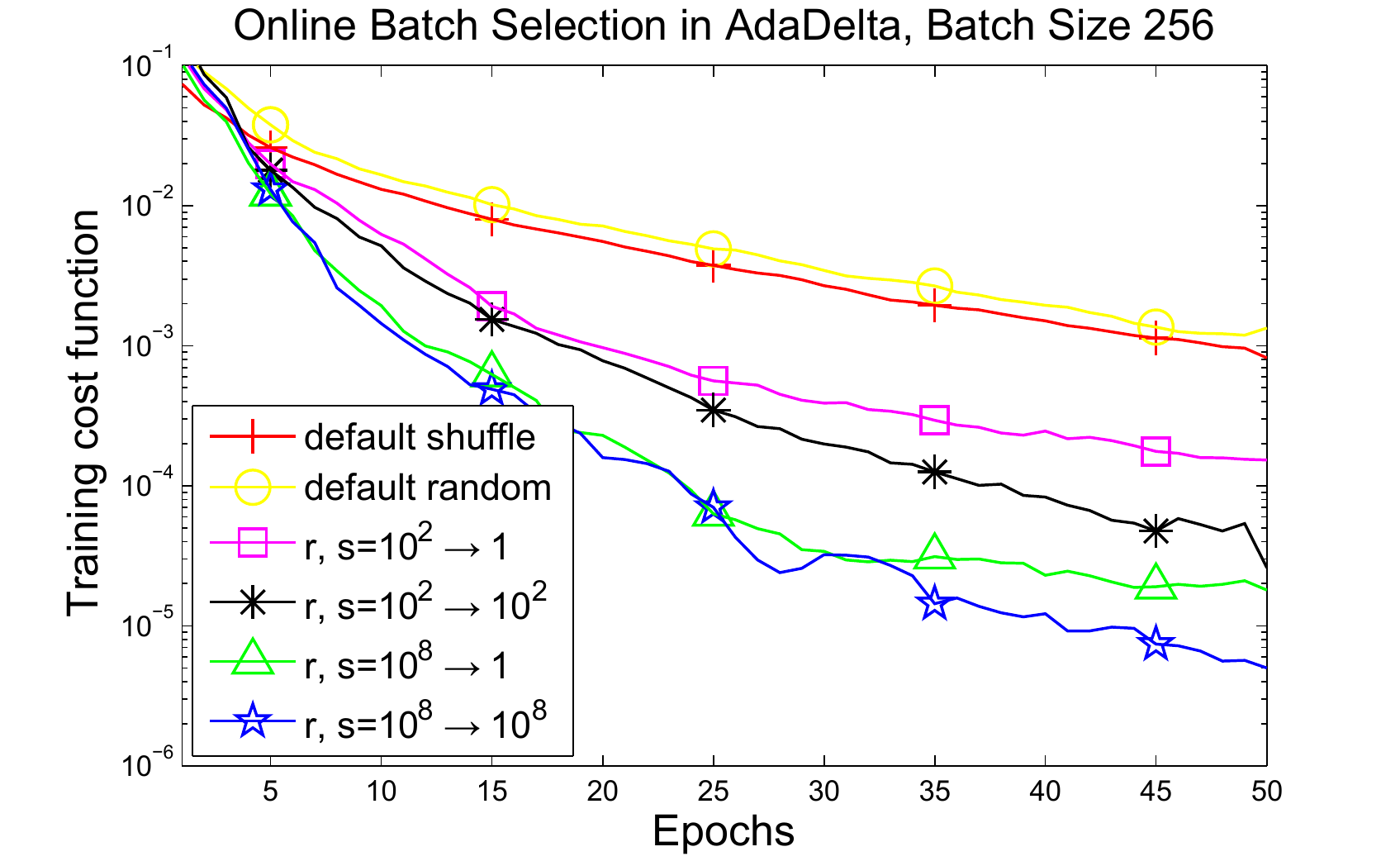}
\includegraphics[width=0.495\textwidth]{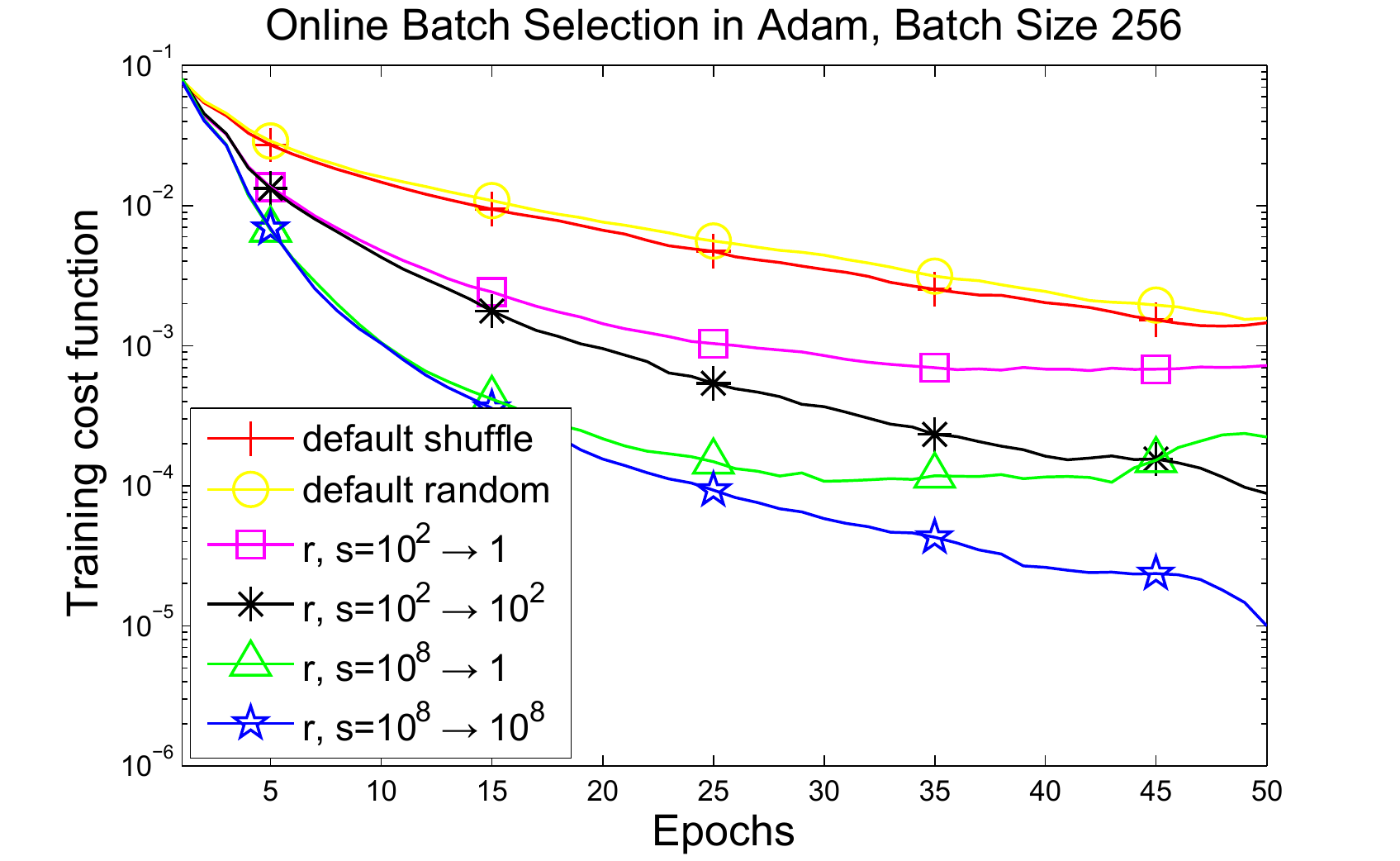}\\
\includegraphics[width=0.495\textwidth]{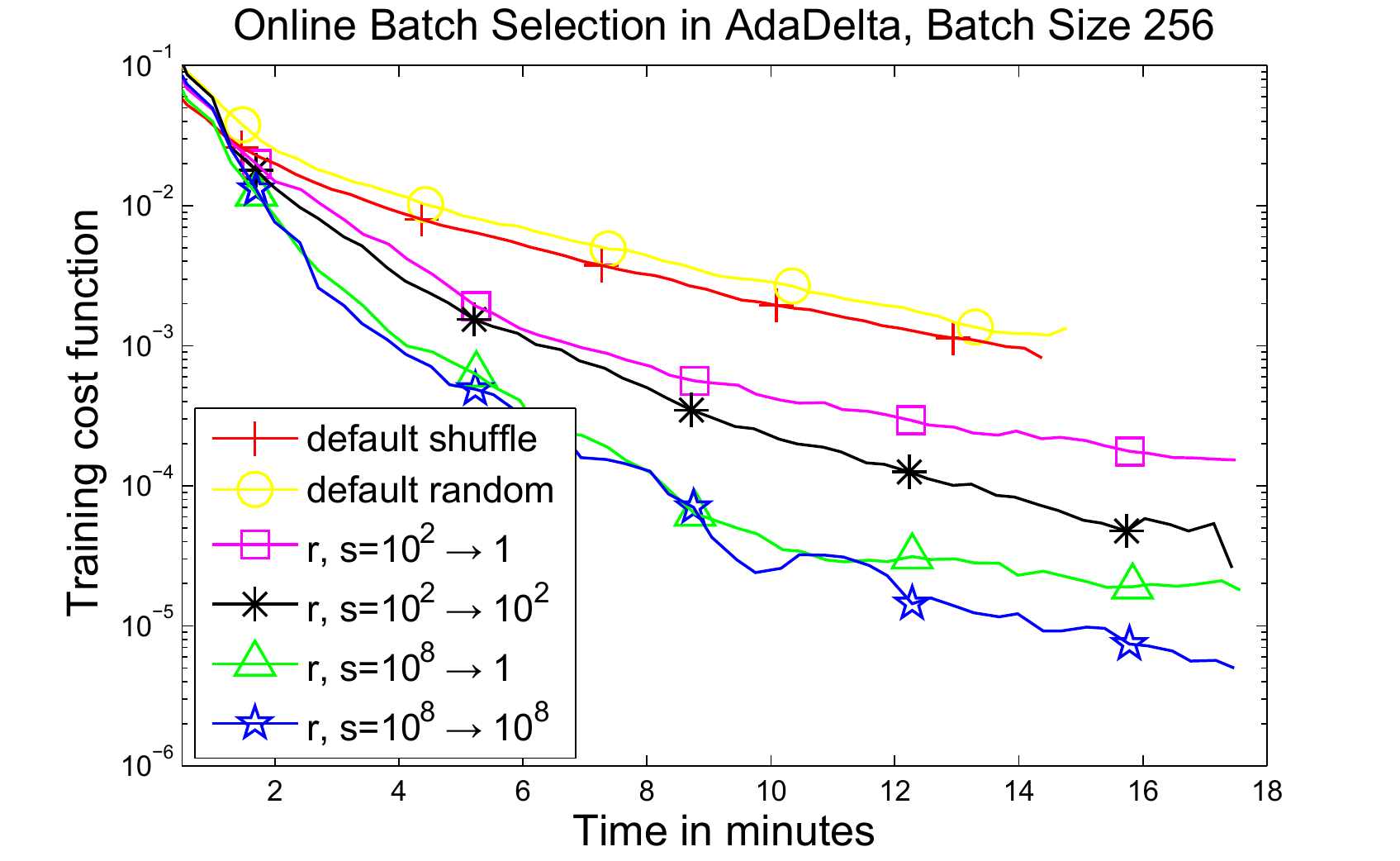}
\includegraphics[width=0.495\textwidth]{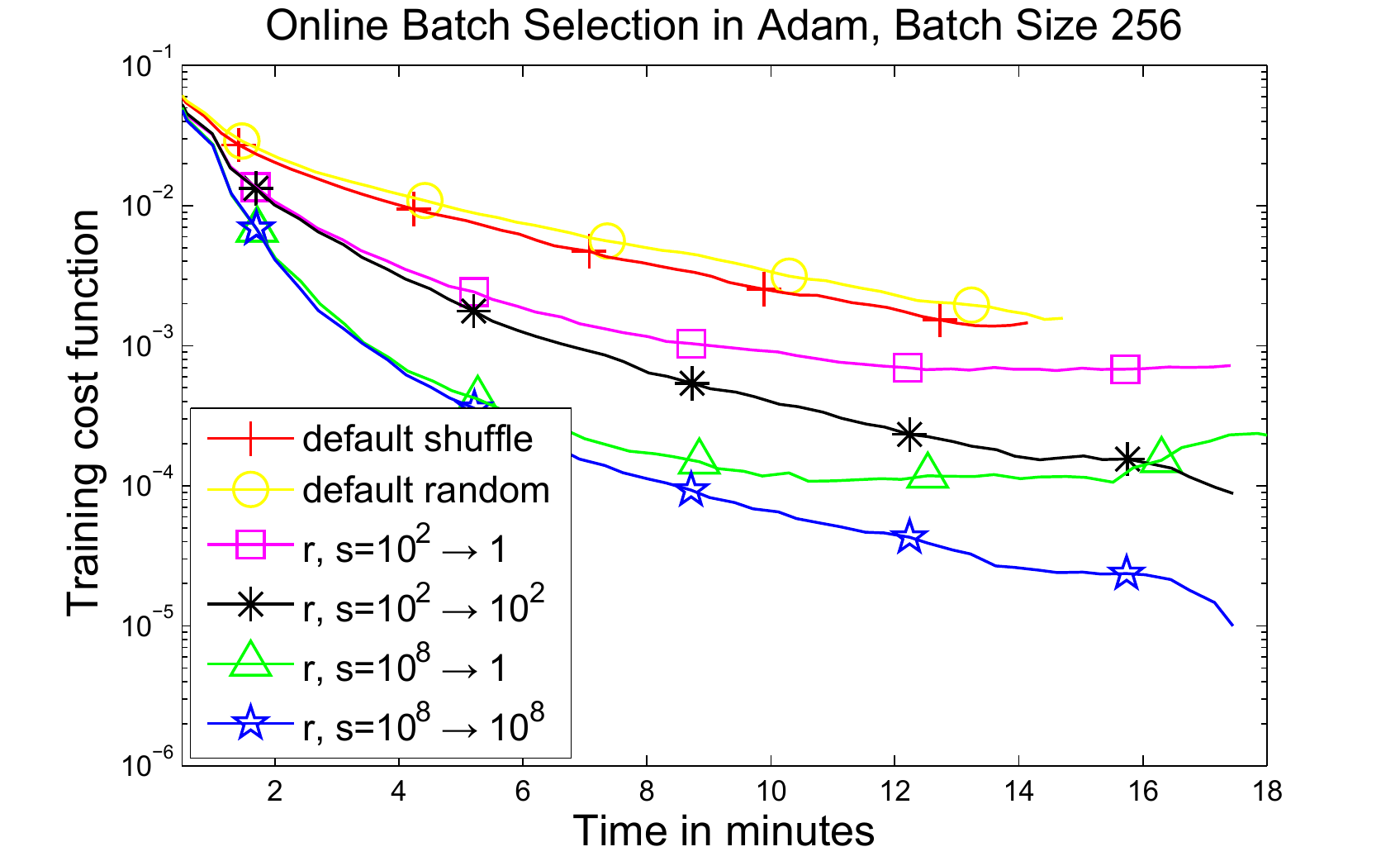}\\
\includegraphics[width=0.495\textwidth]{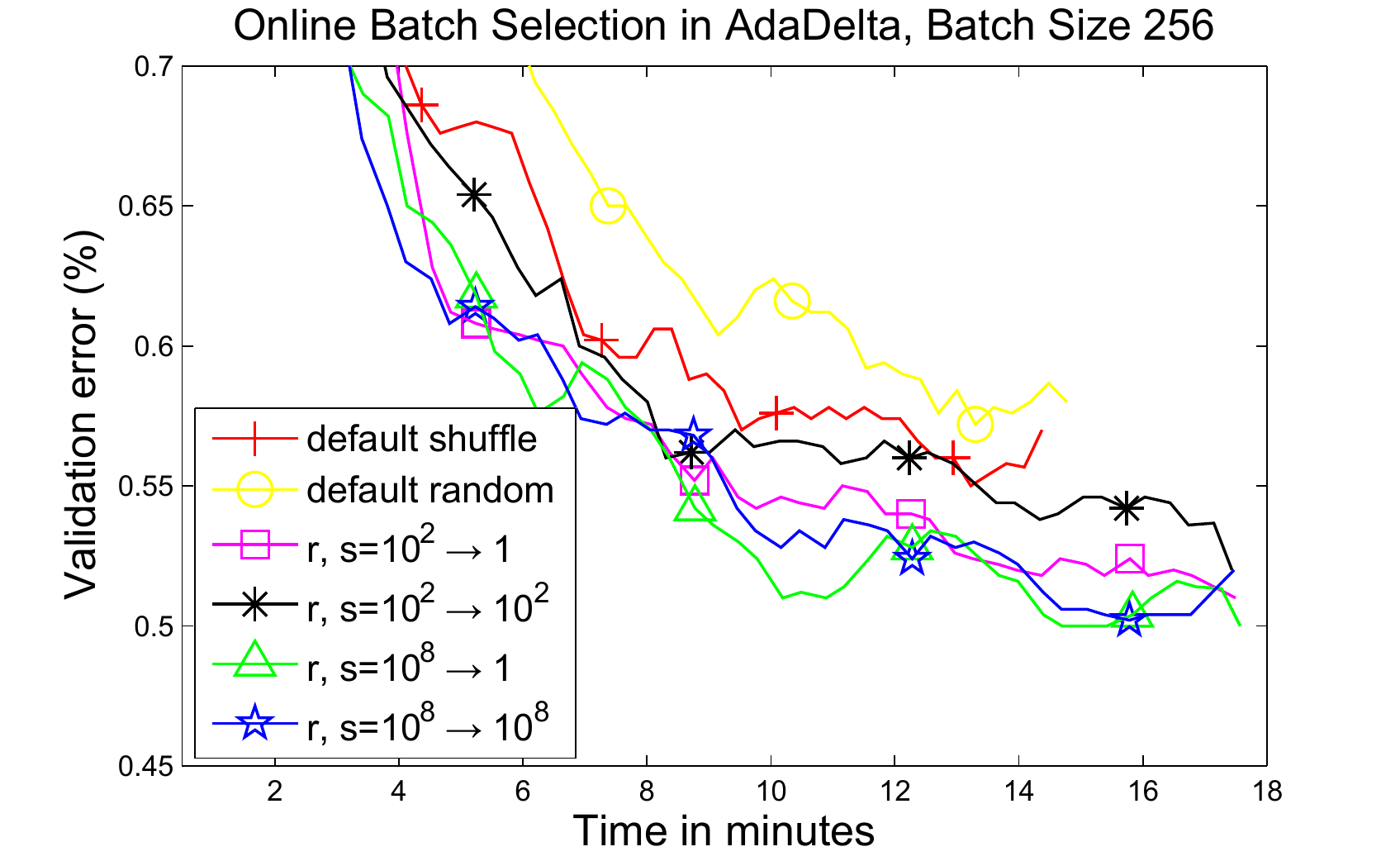}
\includegraphics[width=0.495\textwidth]{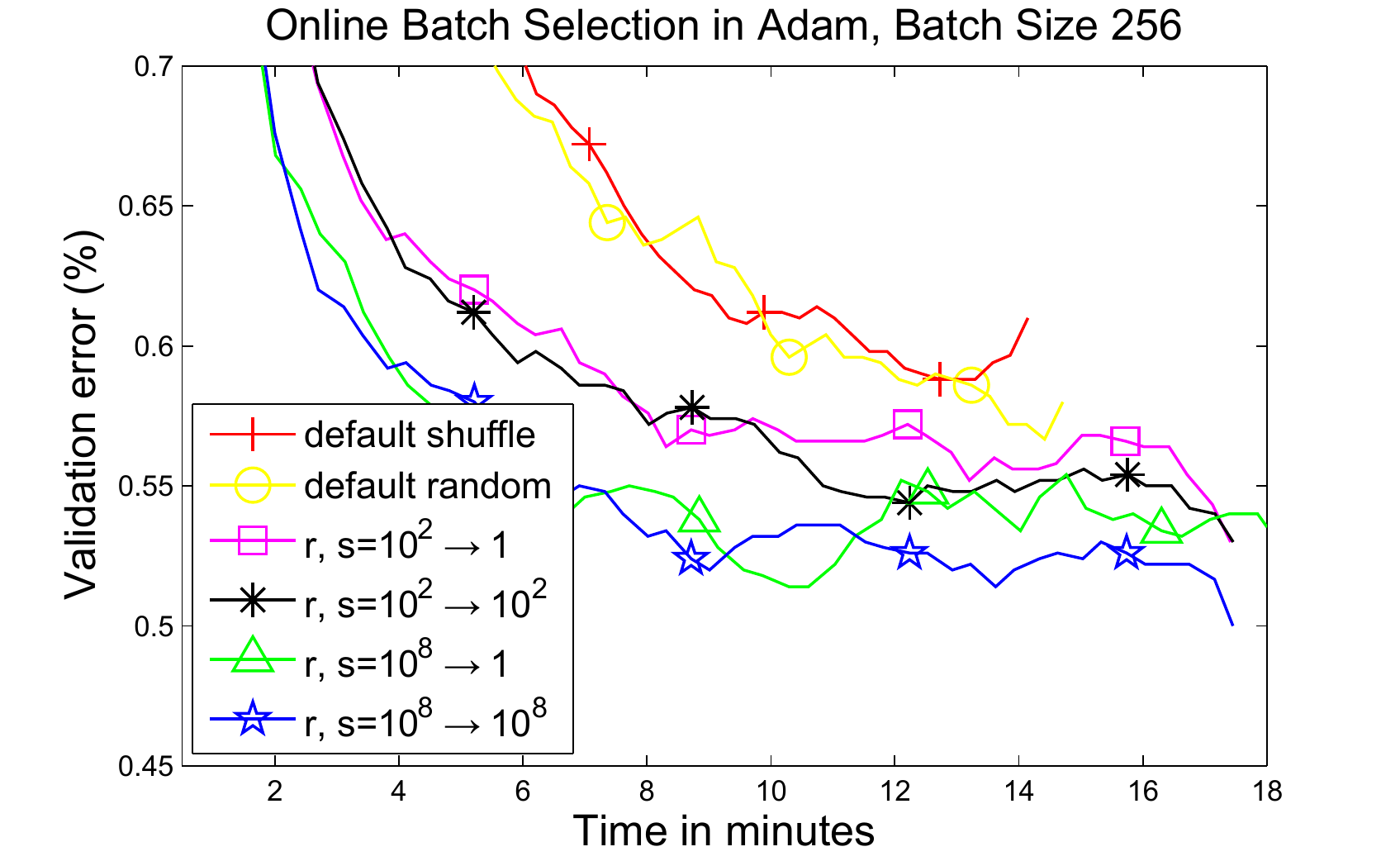}\\
\end{center}
\caption{
Convergence curves of AdaDelta (\textbf{Left Column}) and Adam (\textbf{Right Column}) on MNIST dataset. 
The original algorithms are denoted by ``default shuffle'' (respectively, ``default with random'') when datapoints are shuffled and selected sequentially (respectively, uniformly at random). 
The value of $s$ denotes the ratio of probabilities to select the  training datapoint with the greatest latest known loss rather than the smallest latest known loss. 
Legends with 
$s=s_{e_0} \rightarrow s_{e_{end}}$ correspond to an exponential change of $s$ from $s_{e_0}$ to $s_{e_{end}}$ as a function of epoch index $e$, see Eq. (\ref{eq:ev1}). 
Legends with the prefix ``r'' correspond to the case when the loss values for $r_{ratio} \cdot N=1.0N$ datapoints with the greatest latest known loss 
 are recomputed $r_{freq}=0.5$ times per epoch. 
All curves are computed on the whole training set and smoothed by a moving average of the median of 11 runs.
}
\label{Figure3}
\end{figure*}


\subsection{Hyperparameter optimization}\label{sec:hyperparam_opt}

We investigated whether the use batch selection is still beneficial when hyperparameters are optimized. 
We used Covariance Matrix Adaptation Evolution Strategy (CMA-ES) by \cite{hansen2001completely} in 
its ``active'' regime \citep{hansen2010benchmarking} as implemented in a Python code  
available at \url{https://pypi.python.org/pypi/cma}. 
We used a greater than default population size $\lambda=30$ to benefit from parallel evaluations of solutions on 
a mixed set of 30 GPU's, mostly GeForce GTX TITAN Black. 
Evaluation of each solution involves network building and training using input hyperparameter values. 
The total time for each evaluation is limited. The best validation error found in all epochs represents the objective function.
 Table \ref{table:table} shows the names of hyperparameters, their description and the way to transform the initial 
variables from the ranges $[0,1]$ (note, we constrain the search) to the actual ranges. 
Some variables take different ranges depending on whether we run Adadelta or Adam, 
the others remain inactive (e.g., $x_{1}, x_{2}, x_{5}$) when batch selection is not used. 

\begin{figure*}[p]
\begin{center}
\includegraphics[width=0.495\textwidth]{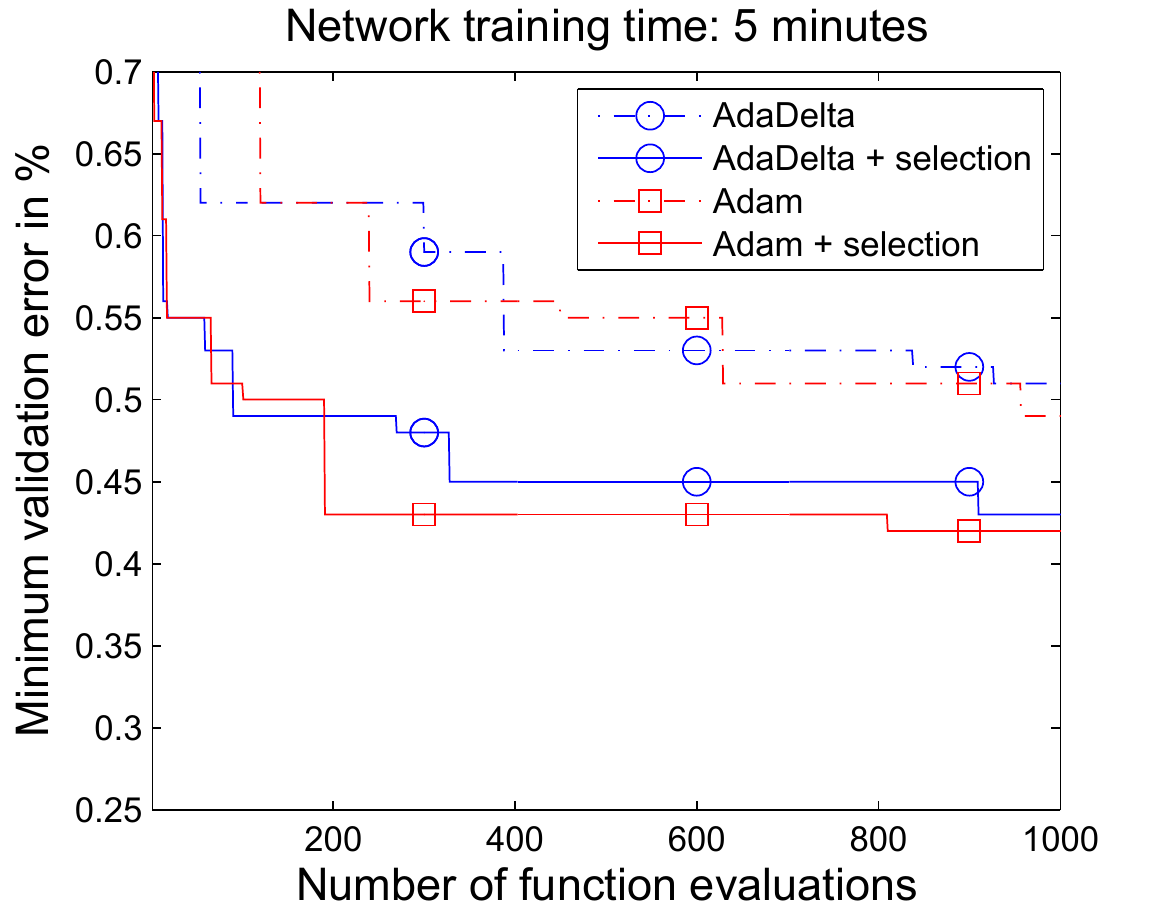}
\includegraphics[width=0.495\textwidth]{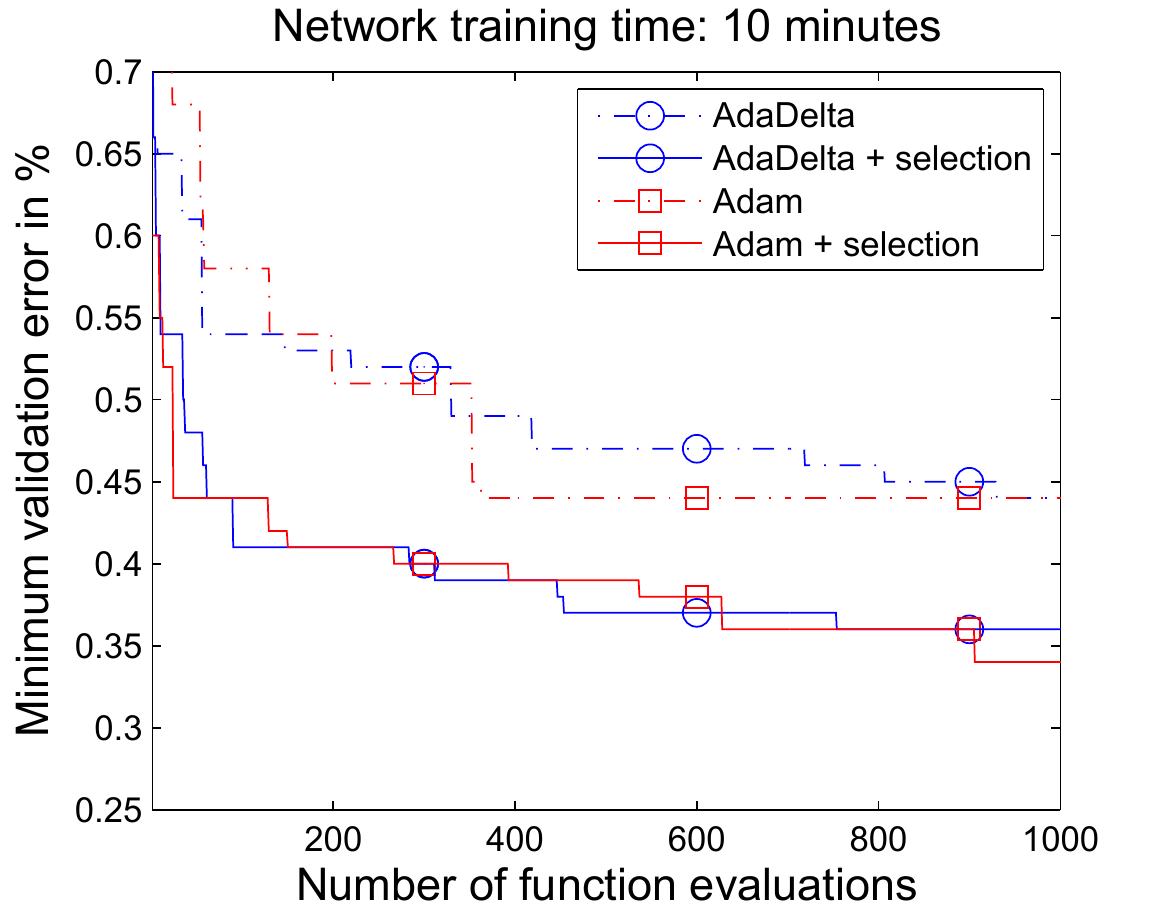}\\
\includegraphics[width=0.495\textwidth]{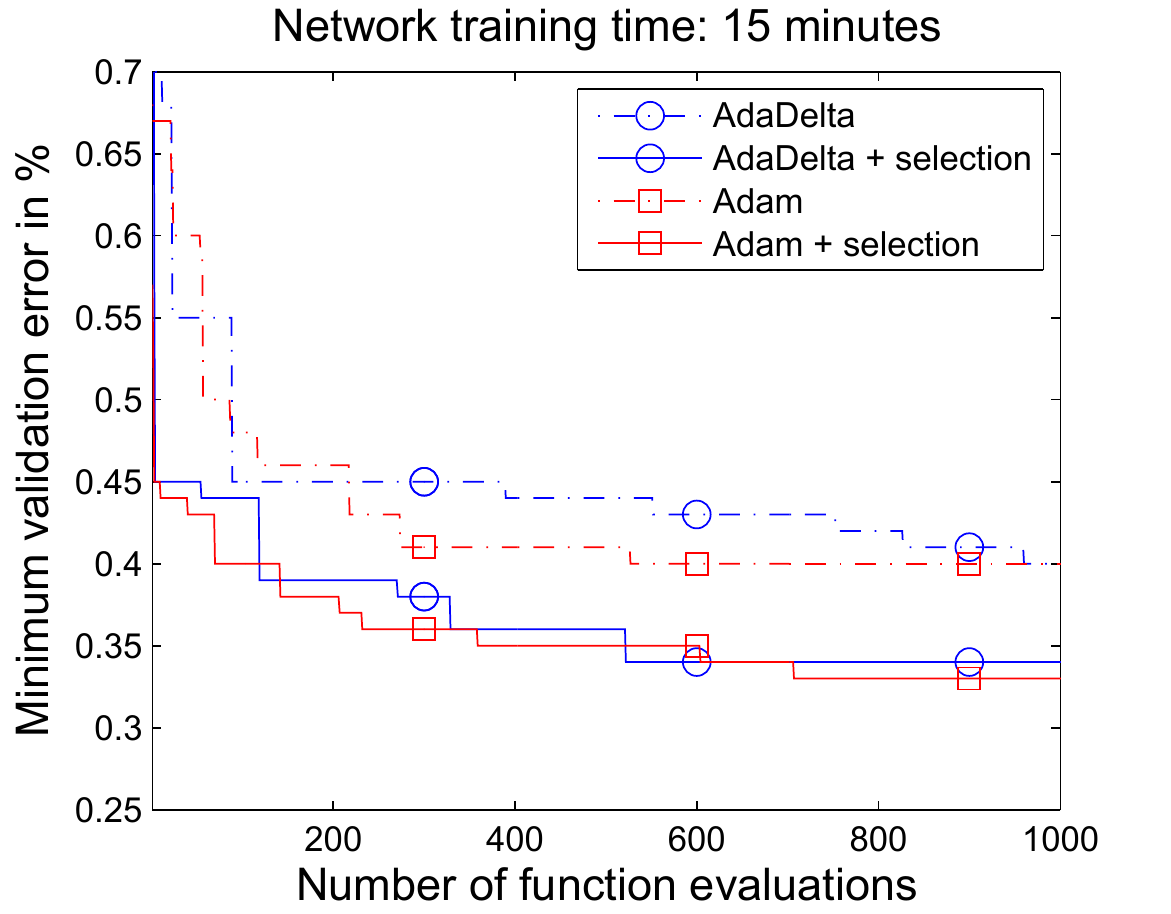}
\includegraphics[width=0.495\textwidth]{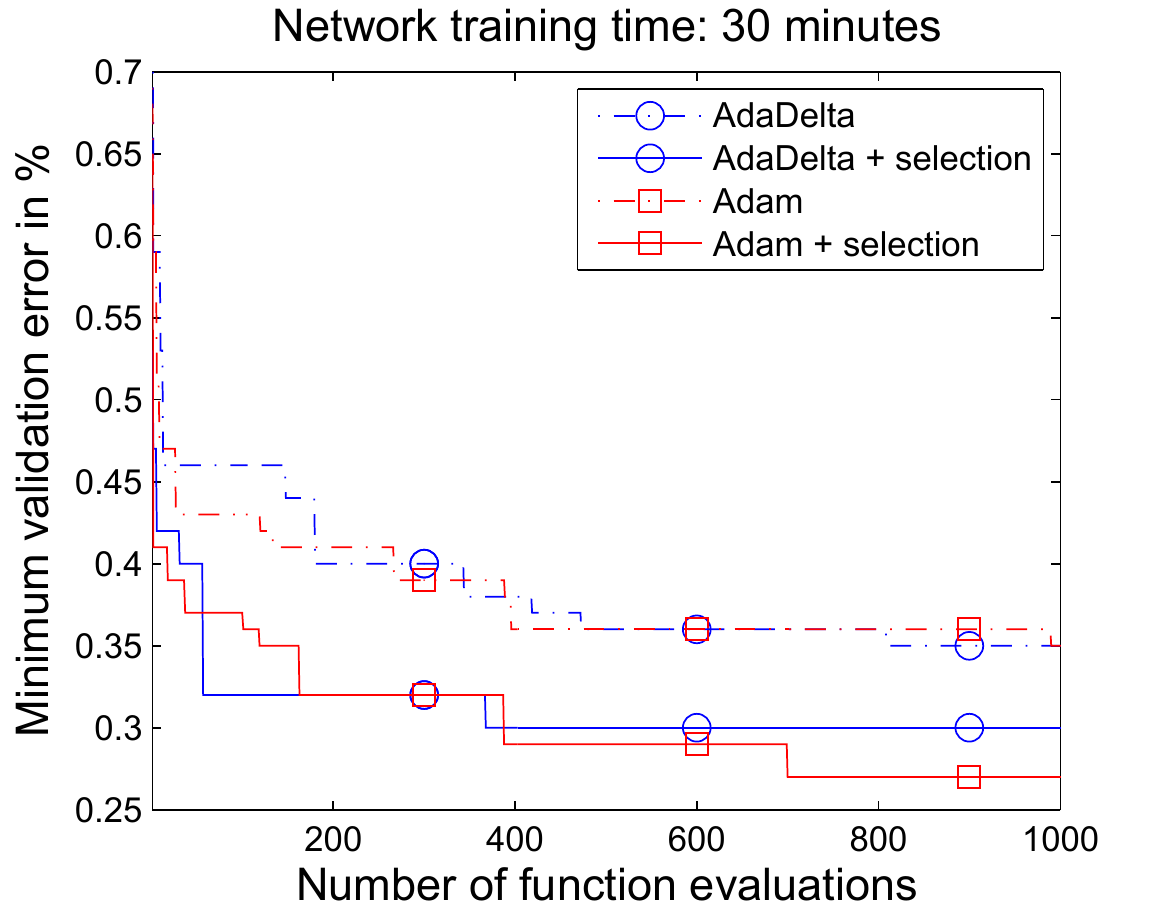}
\end{center}
\caption{
Best validation errors found for AdaDelta and Adam with and without batch selection 
when hyperparameters are optimized by CMA-ES. The results are given for different training time budgets: 
5, 10, 15 and 30 minutes. 
}
\label{FigureMnist}
\end{figure*}

We performed one run for Adadelta, Adam, Adadelta with batch selection and Adam with batch selection 
for time budgets of 5, 10, 15 and 30 minutes. The mean of CMA-ES was set to 0.5 for all variables, the initial step-size was set to 0.2. The results for 34 iterations, about 1000 function evaluations, 
are given in Figure \ref{FigureMnist}. 
The results suggest that batch selection is beneficial both when hyperparameters are random (in the very beginning of the run) 
and optimized (in the end of the run). 
It can be seen that the results found for Adadelta and Adam with batch selection and time budget of 10 minutes are similar 
to the ones found by the original Adadelta and Adam after 30 minutes. 
This suggests that the speedup of a factor of 3 is observed (see also Figure \ref{FigureMnistAll}).

\begin{figure*}[t]
\begin{center}
\includegraphics[width=0.65\textwidth]{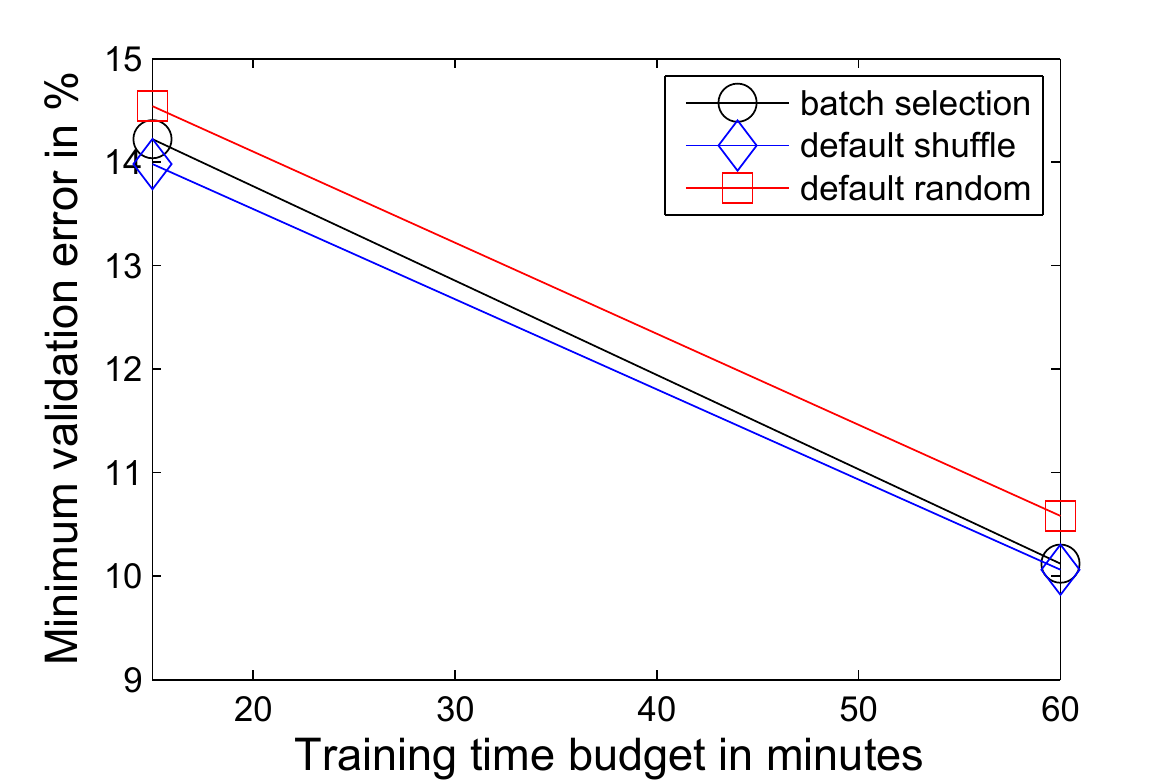}
\end{center}
\caption{
Minimum validation error found on CIFAR-10 for deep neural network trained by Adam with a given budget of time 
(15 and 60 minutes).  
The results correspond to the best hyperparameter settings found by CMA-ES with population size 15 after 510 function evaluations.  
}
\label{FigureCifarAll}
\end{figure*}

Figure \ref{FigureCifarAll} shows the results of our preliminary attempt to apply batch selection for the CIFAR-10 dataset, in a similar way as Figure \ref{FigureMnistAll} in the main paper does for the MNIST dataset.
We note that in this very preliminary experiment on CIFAR-10, batch selection performed slightly better than random selection, but not yet better than shuffling; we are currently analyzing the reasons for this behavior and are extending batch selection to generalize shuffling.
We also note that, for a network training time of 60 minutes, CMA-ES was able to improve validation error of the best solution found from 12\% in the first generation to about 10\% (found at the latest 34-th generation; we used population size 15 on 15 GeForce GTX TITAN X). 

To our best knowledge, our experiments are the first where CMA-ES is used for optimization of 
hyperparameters of deep neural networks. A more detailed analysis and comparison with other techniques is required.

\end{document}